\documentclass[12pt]{article}
\usepackage{PRIMEarxiv}

\usepackage[utf8]{inputenc}
\usepackage[T1]{fontenc}
\usepackage{times}
\usepackage{xspace}
\usepackage{soul}

\usepackage[small]{caption}
\usepackage{wrapfig}
\usepackage{adjustbox}
\usepackage{graphicx}
\graphicspath{{media/}}
\usepackage{dashrule} 
\usepackage{rotating}
\newcommand{\dottedvline}[1]{%
  \rotatebox{90}{\hdashrule[0.5ex]{#1}{0.25pt}{1pt 1pt}}%
}
\usepackage{fix-cm}

\usepackage{amsmath, amssymb, amsfonts, amsthm}
\newtheorem{definition}{Definition}
\usepackage{bm}
\usepackage{nicefrac}

\usepackage[table]{xcolor}

\definecolor{softblue}{RGB}{40, 90, 160}
\definecolor{softred}{RGB}{180, 50, 50}
\definecolor{softmagenta}{RGB}{140, 30, 100}
\definecolor{softgreen}{RGB}{30, 120, 60}

\definecolor{aigold}{RGB}{244,210,1}
\definecolor{aigreen}{RGB}{210,244,211}
\definecolor{aired}{RGB}{255,180,181}
\definecolor{lighterseafoam}{RGB}{194,218,184}
\sethlcolor{aigreen}

\definecolor{LightCyan}{RGB}{232,241,255}
\definecolor{LightRed}{RGB}{255,235,235}
\definecolor{LightPink}{RGB}{255,235,255}
\definecolor{LightGreen}{RGB}{218,255,234}
\definecolor{LightYellow}{RGB}{255,255,235}
\definecolor{LightGray}{RGB}{242,242,242}
\definecolor{Red}{RGB}{253, 239, 242}
\definecolor{Yellow}{RGB}{255, 255, 204}
\definecolor{Pink}{RGB}{255, 243, 254}
\definecolor{Gray}{RGB}{249, 249, 249}
\definecolor{Green}{RGB}{230, 255, 241}
\definecolor{Blue1}{RGB}{218, 232, 245}
\definecolor{Blue2}{RGB}{239, 248, 253}
\definecolor{Blue3}{RGB}{136, 190, 220}
\definecolor{Blue4}{RGB}{83, 157, 204}
\definecolor{Blue5}{RGB}{42, 122, 185}
\definecolor{Blue6}{RGB}{11, 85, 159}
\definecolor{GreenCheck}{RGB}{0, 102, 51}
\definecolor{LightBack}{RGB}{247,249,251}

\usepackage[colorlinks=true,
            linkcolor=cyan, % softred,
            citecolor=softblue,
            urlcolor=softgreen,
            pdfborder={0 0 0}]{hyperref}
\usepackage{enumitem}
\newcommand{\squishlist}{
\begin{list}{{{\small{$\bullet$}}}}{
    \setlength{\itemsep}{1pt}
    \setlength{\parsep}{5pt}
    \setlength{\topsep}{-2pt}
    \setlength{\partopsep}{0pt}
    \setlength{\leftmargin}{2.5em}
    \setlength{\labelwidth}{1em}
    \setlength{\labelsep}{1em}
}}
\newcommand{\squishend}{\end{list}}

\usepackage{booktabs}
\usepackage{colortbl}
\usepackage{array}
\usepackage{tabularx}
\usepackage{makecell}
\usepackage{multirow}
\usepackage{longtable}
\usepackage{arydshln}

\newcolumntype{L}{>{\raggedright\arraybackslash}X}

\usepackage[most, skins]{tcolorbox}
\newtcbtheorem{Summary}{}{enhanced,
  coltitle=black,
  top=0.14in,
  attach boxed title to top left={xshift=0em,yshift=-\tcboxedtitleheight/2},
  boxed title style={size=small,colback=blue!10}
}{summary}

\usepackage{fontawesome5}
\usepackage{pifont}
\usepackage{bbding}

\usepackage{tikz}
\usetikzlibrary{mindmap, shadows}
\tikzset{chatstyle/.style={text width=2.8in, rounded corners=2pt}}

\newcommand{\cmark}{{\color{GreenCheck}\ding{51}}}

\newcommand{\xmark}{{\color{red}\ding{55}}}

\usepackage{algorithm}
\usepackage{algorithmic}
\usepackage{listings}

\usepackage{epigraph}
\usepackage{etoolbox}
\usepackage{xparse}
\usepackage[switch]{lineno}
\usepackage{datetime}
\usepackage{wasysym}

\title{
    Diffusion Models for Low-Light Image Enhancement\\
    \normalsize A Multi-Perspective Taxonomy and Performance Analysis
}

\author{
    {\bfseries Eashan Adhikarla}\quad
    {\bfseries Yixin Liu}\quad
    {\bfseries Brian D. Davison\footnotemark[1]}\quad \\
    \vspace{10pt} \\
    {\bfseries Lehigh University}\quad % {\bfseries $^{1}$Lehigh University}\quad
    \vspace{10pt} \\
    \faGithub~~\url{https://github.com/eashanadhikarla/DMLLIE}\\
}

\begin{document}
\maketitle

\renewcommand{\thefootnote}{\fnsymbol{footnote}}
\footnotetext[1]{Brian D. Davison is the corresponding author: \href{mailto:bdd3@lehigh.edu}{bdd3@lehigh.edu}}
\footnotetext[2]{Latest Update: \today}

\begin{abstract}
        Low-light image enhancement (LLIE) is vital for safety-critical applications such as surveillance, autonomous navigation, and medical imaging, where visibility degradation can impair downstream task performance. Recently, diffusion models have emerged as a promising generative paradigm for LLIE due to their capacity to model complex image distributions via iterative denoising. This survey provides an up-to-date critical analysis of diffusion models for LLIE, distinctively featuring an in-depth comparative performance evaluation against Generative Adversarial Network and Transformer-based state-of-the-art methods, a thorough examination of practical deployment challenges, and a forward-looking perspective on the role of emerging paradigms like foundation models. We propose a multi-perspective taxonomy encompassing six categories: Intrinsic Decomposition, Spectral \& Latent, Accelerated, Guided, Multimodal, and Autonomous; that map enhancement methods across physical priors, conditioning schemes, and computational efficiency. Our taxonomy is grounded in a hybrid view of both the model mechanism and the conditioning signals. We evaluate qualitative failure modes, benchmark inconsistencies, and trade-offs between interpretability, generalization, and inference efficiency. We also discuss real-world deployment constraints (\textit{e.g.}, memory, energy use) and ethical considerations. This survey aims to guide the next generation of diffusion-based LLIE research by highlighting trends and surfacing open research questions, including novel conditioning, real-time adaptation, and the potential of foundation models.
\end{abstract}

\newpage
\tableofcontents
\newpage

%%%%%%%%%%%%%%%%%%%%%%
\section{Introduction}\label{sec:introduction}
%%%%%%%%%%%%%%%%%%%%%%
    Low-light image enhancement (LLIE) is a foundational problem in computer vision, vital for domains such as surveillance, autonomous driving, medical diagnostics, and consumer photography. The increasing demand for robust LLIE in an ever-expanding range of applications, particularly those requiring high reliability in uncontrolled environments like autonomous systems operating at night, underscores the urgency of advancing this field. Inadequate illumination causes significant degradation in image quality, manifested as low contrast, signal-dependent noise, and suppressed textures, which in turn reduces the reliability of downstream tasks like object detection ~\cite{guo2021dynamic,jiang2025low}, recognition ~\cite{gong2025optimization,li2025ldwle}, and segmentation ~\cite{tan2025darksegnet}. Unlike denoising or deblurring, LLIE must solve a fundamentally ill-posed inverse problem: reconstructing semantically meaningful structure from severely underexposed, often ambiguous observations. Traditional enhancement methods such as histogram equalization~\cite{celik2011contextual,lee2013contrast}, gamma correction~\cite{wang2023low,severoglu2025acgc}, and traditional Retinex-based approaches~\cite{9229994,traditionalretinex} depend heavily on handcrafted priors or global adjustments. These methods often fail under complex lighting conditions, suffer from over-enhancement or color distortions \cite{URetinexnet,zamir2022restormer,Cui_2022_BMVC}, and lack adaptability to real-world scenarios, particularly in safety-critical applications where incorrect enhancement could lead to false interpretations or decisions.

    Originally developed for generative modeling~\cite{ho2020ddpm}, diffusion models have recently gained traction in low-level vision tasks such as LLIE~\cite{he2024diffusion}, owing to their robustness and flexibility. The surge in popularity of diffusion models for LLIE stems from their remarkable ability to generate highly realistic details and textures, addressing a key failing of earlier methods. Their probabilistic nature, simulating a forward noising process and learning the reverse denoising trajectory, typically via score matching or solving stochastic differential equations (SDEs), is particularly well-suited for the ill-posed nature of LLIE. This framework treats enhancement as an inverse problem, recovering high-quality images from heavily degraded inputs by modeling the underlying data distribution. Unlike Generative Adversarial Network (GAN)-based methods (\textit{e.g.}, PD-GAN~\cite{liu2021pd}, LE-GAN~\cite{fu2022gan}, MFGAN~\cite{jung2020multi}), diffusion models offer greater training stability, mitigate mode collapse, and produce more reliable results under extreme lighting conditions. They also allow for explicit conditioning on physical priors (\textit{e.g.}, exposure, entropy), enabling unsupervised learning even in the absence of paired datasets (see Section~\ref{subsec:adaptive}). The field of LLIE is increasingly viewing the problem not just as restoring a single ground truth, but as generating a visually plausible and detailed image consistent with the low-light input, a paradigm for which diffusion models are inherently suited due to their capacity to model complex distributions and learn strong image priors.
    The landscape of generative models for LLIE can be viewed through the lens of a ``generative trilemma,''~\cite{xiao2022DDGAN} where diffusion models, GANs, and transformers each present a unique set of trade-offs concerning sample quality/realism, training stability/mode collapse, and computational efficiency/inference speed~\cite{wolf2025diffusionmodelsroboticmanipulation}. 
    Diffusion models represent a significant point in this trade-off space, typically excelling in output quality and training stability but initially facing challenges with inference speed, a limitation that is now being actively addressed by the research community, as will be discussed in Section~\ref{subsec:efficiency}.

    \begin{figure*}[ht]
        \centering
        \includegraphics[width=0.85\linewidth]{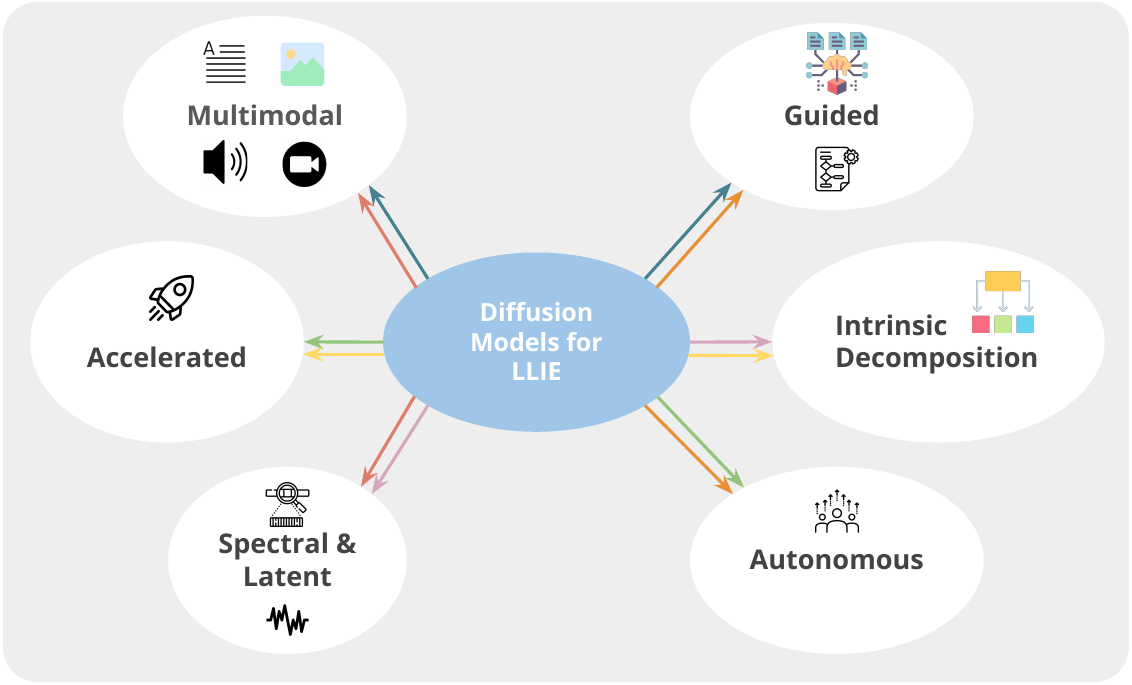}
        \caption{\label{fig:teaser}Categorical representation of widely used diffusion methods in low-light into six different categories (Best viewed in color.)}
    \end{figure*}

    As the field moves toward more interpretable, controllable, and generalizable enhancement pipelines, diffusion models have become central to next-generation LLIE research. Recent diffusion-based LLIE frameworks have diversified along multiple methodological axes, including how they handle degradation priors, where they operate (image, frequency, or latent space), and the types of conditioning they allow. To structure this growing body of work, we propose a six-category taxonomy (illustrated in Figure~\ref{fig:teaser}): (\textbf{1}) Intrinsic Decomposition, (\textbf{2}) Spectral \& Latent, (\textbf{3}) Accelerated, (\textbf{4}) Guided, (\textbf{5}) Multimodal, and \textbf{(6)} Autonomous diffusion models. This taxonomy captures distinctions in enhancement strategy, such as physically grounded decomposition~\cite{wang2023lldiffusion}, domain-transformed processing (\textit{e.g.}, latent~\cite{jiang2023low}, Fourier~\cite{L2DM}, wavelet~\cite{wan2024sfdiff}), user or task-driven guidance, and self-supervised or zero-shot adaptation. While several surveys have touched upon diffusion models in broader low-level vision (\textit{e.g.}, He et al.~\cite{he2024diffusion}) or specific aspects of LLIE, this work offers a most recent synthesis focused specifically on diffusion-based LLIE. This survey contributes: (\textbf{1}) a refined and critically analyzed taxonomy reflecting the latest methodological trends; (\textbf{2}) a broad quantitative and qualitative comparison, benchmarking diffusion models against leading GAN and Transformer architectures on diverse datasets; (\textbf{3}) an in-depth dissection of persistent challenges: analyzing interdependencies among challenges and their implications for deployment; and (\textbf{4}) a forward-looking analysis of emerging research frontiers, including the integration of foundation models and the drive towards on-device solutions. While many methods share overlapping features, our categorization highlights their primary innovation axis and reflects broader trends: toward real-time efficiency, controllability, and generalization to unpaired data. This survey systematically analyzes representative works under each category, compares their assumptions, strengths, and limitations, and outlines emerging cross-domain synergies (\textit{\textit{e.g.}}, with dehazing or text recognition). In doing so, we aim not only to synthesize recent progress, but to surface core research challenges and provide a roadmap for future work in diffusion-based LLIE.

    The remainder of this work is organized as follows: First, in Section \ref{sec:insights} we share some observations and insights based on the extensive diffusion model literature, which set the stage for the more detailed review that follows.  Section~\ref{sec:background} provides background on LLIE challenges and diffusion model fundamentals. Section~\ref{sec:taxonomy} details our proposed taxonomy of diffusion models for LLIE. Section~\ref{sec:datasets-metrics} discusses benchmark datasets, evaluation metrics, and presents a comprehensive performance analysis. Section~\ref{sec:challenges_limitations} delves into the ongoing challenges and limitations. Section~\ref{sec:future} explores future research directions. Finally, Section~\ref{sec:conclusion} concludes the survey.

%%%%%%%%%%%%%%%%%%%%%%%%%%%%%%%%%%%%%%%%%%%%%
\section{Key Observations and Insights} 
\label{sec:insights}
%%%%%%%%%%%%%%%%%%%%%%%%%%%%%%%%%%%%%%%%%%%%%

    To facilitate an overall understanding of our survey, this section synthesizes the critical observations and insights drawn from the extensive body of work on diffusion models for LLIE. Our analysis reveals that the field is not evolving along a single axis of improvement but is instead navigating a series of fundamental trade-offs.

\begin{figure}
        \centering
        \includegraphics[width=0.9\linewidth]{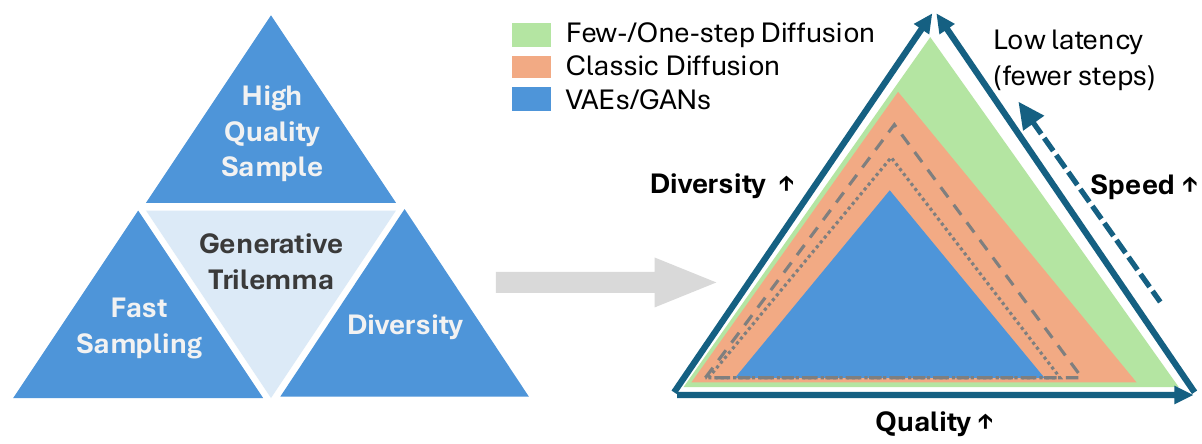}
        \caption{\label{fig:trilemma}\textbf{Expanding feasible frontier in LLIE.} Axes: quality$\uparrow$, diversity$\uparrow$; interior = lower latency. Envelopes for VAEs/GANs (2017--2020, blue), classic diffusion (2021--2023, orange), and few/one-step/consistency diffusion (2023--2025, green) illustrate a monotonic expansion.}
    \end{figure}

    \textbf{The LLIE Generative Trilemma.} Early LLIE generative models appeared to face a three-way tension among quality, diversity, and latency. Over the 2021–2025 period, diffusion models, augmented by distillation, rectified/consistency flows, and latent-space operations, have expanded the feasible region rather than merely sliding along a fixed trade-off. In practice, modern systems attain higher perceptual quality and better mode coverage at substantially lower sampling cost than prior VAE/GAN baselines. We visualize this shift in Figure~\ref{fig:trilemma}: the green region (2023–2025) strictly contains the orange (2021–2023) and blue (2017–2020), indicating simultaneous gains across axes. This expansion is not ``free'', it is powered by techniques that recur throughout our taxonomy: accelerated diffusion (distillation, step-truncation), spectral/latent formulations (Fourier/wavelet/latent spaces), and guided or task-aware conditioning that reduces wasted compute on irrelevant modes. The remaining bottlenecks have moved from sampling steps vs. fidelity to memory footprint, on-device constraints, and controllability under distribution shift, themes we revisit in Sections~\ref{sec:taxonomy}–\ref{sec:challenges_limitations}.

    \textbf{The ``Priors vs. Plasticity'' Dilemma.} There is a fundamental tension between embedding strong physical priors and maintaining model flexibility (plasticity). Intrinsic Decomposition methods, grounded in Retinex theory or physics-driven models, offer high interpretability and can prevent a model from generating physically implausible results. However, they are constrained by the accuracy of their underlying assumptions, which can fail in scenes with complex, non-uniform lighting.

    In contrast, autonomous diffusion methods that rely on zero-shot or self-supervised learning  offer superior generalization to unseen lighting conditions by not being tied to a specific physical model. The trade-off is a potential loss of fidelity and the risk of ``hallucinating'' details that are plausible but not faithful to the original scene content.

    \textbf{A Paradigm Shift from Restoration to Controllability.} The focus of LLIE is evolving from a simple, monolithic restoration task to a more nuanced, controllable, and context-aware process. Early deep learning methods aimed to find a single ``best'' enhancement for a given input. The rise of diffusion models has enabled a more flexible paradigm:
    \begin{itemize}[leftmargin=8pt,itemsep=0.5pt]
        \item Guided Diffusion methods (Section \ref{subsec:adaptive}) allow for explicit control via region-based masks, user instructions, or exposure parameters. This transforms LLIE from an automatic process into an interactive one.
        \item Multimodal and Task-Specific Diffusion methods (Section \ref{subsec:specialized}) further this trend by optimizing the enhancement not for generic perceptual quality, but for the success of a specific downstream task, such as text recognition or object detection, or by leveraging alternate sensor data like event cameras. This shift reflects a move from single-output restoration to controllable, task-aware enhancement, where ``enhancement'' is defined by its utility, not just its appearance.
    \end{itemize}

%%%%%%%%%%%%%%%%%%%%%%%%%%%%%%%%%%%%%%%%%%%%%
\section{Background} % includes related work
%%%%%%%%%%%%%%%%%%%%%%%%%%%%%%%%%%%%%%%%%%%%%
\label{sec:background}
    \subsection{Core Challenges in Low-Light Image Enhancement}
    
    Effectively enhancing low-light images requires addressing a complex set of interrelated degradation factors that arise from capturing scenes with insufficient photons. These challenges are not isolated but often compound each other; for instance, attempts to brighten an image inevitably amplify existing noise, and lost details become even harder to plausibly generate if color and contrast cues are also weak. Understanding these challenges is crucial for appreciating the design choices and limitations of enhancement algorithms, including those based on diffusion models.
    
    \textbf{Noise}: Low-light images are notoriously noisy~\cite{zheng2022low}. The low photon count leads to a low signal-to-noise ratio (SNR)~\cite{huang2024detail}. The noise is often a complex mixture of shot noise (related to photon arrival statistics) and various types of sensor noise (read noise, thermal noise). Crucially, this noise is often signal-dependent and non-Gaussian, deviating significantly from the additive white Gaussian noise assumption used in many standard image restoration models~\cite{hou2023global}. Processing RAW sensor data can provide access to more linear noise characteristics but introduces challenges related to demosaicing and color processing. Diffusion models operating on or informed by RAW data (\textit{e.g.}, as used with the SID dataset~\cite{Chen2018SID}) can be particularly effective as RAW data bypasses some in-camera processing artifacts that can complicate noise modeling. A major difficulty is that simply increasing the brightness of a low-light image inevitably amplifies the existing noise, often making the enhanced image visually unacceptable~\cite{zheng2022low}. Therefore, effective LLIE methods must perform simultaneous brightening and denoising.
    
    \textbf{Low Contrast and Visibility}: Insufficient illumination leads to a compressed dynamic range, resulting in low contrast and poor visibility of scene details, especially in dark or shadowed regions~\cite{zheng2022low}. The problem is often exacerbated by non-uniform illumination within the scene, where some areas might be adequately lit while others are in deep shadow, requiring spatially adaptive enhancement~\cite{cmc.2025.058495}.
    
    \textbf{Color Distortion}: Low-light significantly affects the accuracy of color capture. Images may exhibit color casts (\textit{e.g.}, a shift towards blue or yellow), reduced saturation, or inaccurate color reproduction~\cite{9229994}.
    Correcting these color issues while enhancing brightness without introducing new artifacts is a significant challenge~\cite{cmc.2025.058495}.
    
    \textbf{Detail/Texture Loss}: Perhaps the most fundamental challenge is the loss of fine-grained details and textures, particularly in regions plunged into extreme darkness~\cite{huang2024detail}.
    The information might be severely degraded or entirely absent in the captured signal. Recovering these details often requires algorithms to infer or generate plausible textures based on context and learned priors, making LLIE an ill-posed problem~\cite{zheng2022low}.
    
    \textbf{Data Scarcity and Generalization}: Training deep learning models for LLIE ideally requires large datasets of perfectly aligned image pairs captured under identical scene conditions but varying illumination levels (low-light and normal light)~\cite{zheng2022low}. Acquiring such data, especially for dynamic scenes, is extremely difficult and costly due to challenges in ensuring perfect spatial alignment and identical scene content across different exposures~\cite{zheng2022low}. Existing datasets often have limitations in terms of size, diversity of scenes, resolution, or capture conditions~\cite{zheng2022low}. 
    This data scarcity poses a major obstacle for supervised learning methods, often limiting their ability to generalize well to unseen real-world low-light images that may exhibit different noise characteristics, lighting patterns, or scene content than the training data~\cite{zheng2022low}.
    This ``chicken-and-egg'' dilemma, where data scarcity hinders robust model development, which in turn makes it difficult to demonstrate utility and secure resources for better data collection, particularly fuels research into unsupervised, zero-shot, and self-supervised diffusion models (discussed in Section~\ref{subsec:adaptive})~\cite{lightendiffusion,yang2023difflle}.
    These alternative learning paradigms, however, introduce their own challenges related to achieving high fidelity and controlled generation without strong explicit priors. 
    This limitation motivates the exploration of unsupervised, zero-shot, or semi-supervised learning paradigms, as well as the use of generative models with strong priors~\cite{lin2024aglldiff,fu2022gan}.
    Furthermore, a ``good'' enhancement is not solely defined by human visual appeal. For many practical systems, the utility of LLIE is measured by its impact on downstream machine vision tasks, such as object detection or semantic segmentation. These tasks may have different optimal enhancement characteristics than what a human observer prefers, motivating the development of task-specific LLIE approaches, as explored in Section~\ref{subsec:specialized}.
    
    The confluence of these factors, including complex noise, low contrast, color shifts, severe detail loss, and data limitations, makes LLIE a particularly demanding task. It necessitates algorithms that can simultaneously enhance brightness, suppress complex noise, restore color fidelity, and recover plausible details, often with limited or imperfect training data. This complex interplay motivates the investigation of powerful generative approaches like diffusion models, which offer the potential to learn strong priors about natural images and generate realistic details, alongside unsupervised or zero-shot learning strategies to mitigate data dependency.

    \textbf{Historical Context of Traditional Enhancement Pipelines.} Early approaches to LLIE were predominantly built around handcrafted priors and deterministic transformations, such as histogram equalization (HE), gamma correction, and Retinex-based decomposition methods~\cite{traditionalretinex,9229994}. These methods aimed to manipulate intensity or illumination estimates but often amplified noise or produced unnatural tones under complex lighting. As deep learning gained traction, methods like EnlightenGAN~\cite{fu2022gan}, Zero-DCE~\cite{guo2020zero}, and LLformer~\cite{wang2023ultra} emerged, introducing end-to-end learning strategies for enhancement and denoising~\cite{fu2022gan}. Some models incorporated Retinex priors into CNN architectures (\textit{e.g.}, KinD, SCI), while others pursued generative modeling via GANs or invertible flows. Despite improvements in detail recovery and perceptual quality, these methods frequently suffered from mode collapse, unstable training, and poor generalization under extreme degradations or out-of-distribution lighting conditions~\cite{dhariwal2021diffusion}.
    These limitations laid the groundwork for the current paradigm shift toward diffusion models, which offer stable training, flexible conditioning, and a natural probabilistic formulation suited for ill-posed restoration tasks.

    \subsection{Fundamentals of Diffusion Models}

    Diffusion models, including Denoising Diffusion Probabilistic Models (DDPMs)~\cite{ho2020ddpm}, are built upon a two-phase stochastic formulation, namely a forward process that perturbs clean data by incrementally adding Gaussian noise, and a learned reverse process that restores samples through iterative denoising. Their recent success in image synthesis, editing, and restoration tasks stems from their ability to model complex data distributions with fine-grained control over generation. Low-light image enhancement (LLIE), often an ill-posed inverse problem, benefits from diffusion’s iterative and uncertainty-aware nature, which allows flexible conditioning, unsupervised learning, and incorporation of domain priors.

    \begin{figure}[ht!]
        \centering
        \includegraphics[width=0.8\linewidth]{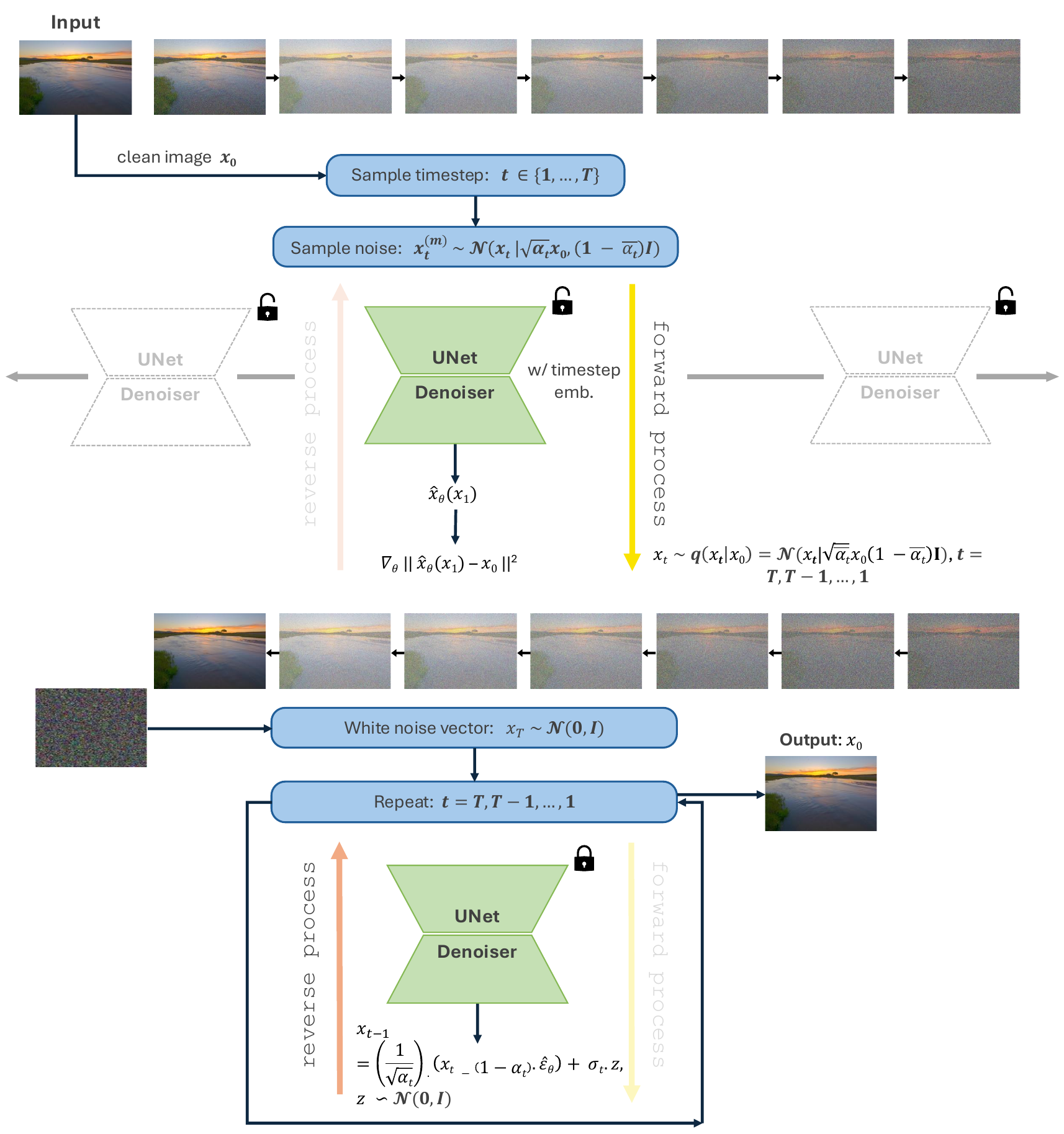}
        \caption{\label{fig:diffusion-pipeline}Illustration of the diffusion training and inference pipeline. \textbf{Top:} The forward process gradually perturbs the clean image $x_0$ into noisy representations $x_t$ by adding Gaussian noise across timesteps. \textbf{Bottom:} The reverse process starts from white noise $x_T \sim \mathcal{N}(0,I)$ and employs a U-Net denoiser $\epsilon_\theta(x_t, t)$ to iteratively recover the clean image. Training optimizes the noise prediction loss $\mathcal{L}_{\text{simple}}$, while inference repeatedly applies the reverse step to reconstruct $x_0$.}
    \end{figure}

    \subsubsection{Forward and Reverse Process.}
    Diffusion models are built on a two-phase stochastic framework that is well illustrated in the training and inference pipeline (Figure~\ref{fig:diffusion-pipeline}). The \textbf{forward process} gradually corrupts a clean image $x_0 \sim q(x_0)$ into a noisy sequence $\{x_t\}_{t=1}^T$ by adding Gaussian noise in incremental steps. At each timestep $t$, the noisy sample is generated as:
    \begin{equation}
        q(x_t \mid x_{t-1}) = \mathcal{N}\!\bigl(x_t; \sqrt{1-\beta_t}\, x_{t-1}, \, \beta_t I \bigr),
    \end{equation}
    where $\{\beta_t\}_{t=1}^T$ is the variance schedule controlling the signal-to-noise ratio. By marginalizing, one can directly obtain:
    \begin{equation}\label{eq:forward}
        q(x_t \mid x_0) = \mathcal{N}\!\bigl(x_t; \sqrt{\bar{\alpha}_t}\, x_0, (1-\bar{\alpha}_t)I \bigr), \quad 
        \bar{\alpha}_t = \prod_{s=1}^t (1-\beta_s).
    \end{equation}
    
    Thus, $x_t$ can be efficiently sampled from $x_0$ by combining it with Gaussian noise $\epsilon \sim \mathcal{N}(0,I)$:
    \begin{equation}
        x_t = \sqrt{\bar{\alpha}_t}\, x_0 + \sqrt{1-\bar{\alpha}_t}\,\epsilon.
    \end{equation}
    
    \noindent
    The \textbf{reverse process} learns to iteratively denoise starting from a pure Gaussian noise vector $x_T \sim \mathcal{N}(0,I)$, reconstructing an estimate of $x_0$. A U-Net denoiser $\epsilon_\theta(x_t, t)$ is trained to predict the noise component, enabling the parameterization:
    \begin{equation}
        p_\theta(x_{t-1} \mid x_t) = \mathcal{N}\!\bigl(x_{t-1}; \mu_\theta(x_t, t), \, \Sigma_\theta(x_t, t)\bigr),
    \end{equation}
    with mean expressed as:
    \begin{equation}
        \mu_\theta(x_t, t) = \frac{1}{\sqrt{\alpha_t}}\Bigl(x_t - \frac{1-\alpha_t}{\sqrt{1-\bar{\alpha}_t}} \, \epsilon_\theta(x_t, t)\Bigr).
    \end{equation}
    
    During inference, this denoising step is repeated from $t=T$ down to $t=1$, progressively refining the sample toward a clean image. In practice, stochasticity is preserved by injecting Gaussian noise $\mathbf{z} \sim \mathcal{N}(0,I)$ at each step:
    \begin{equation}
        x_{t-1} = \mu_\theta(x_t, t) + \sigma_t \mathbf{z},
    \end{equation}
    where $\sigma_t$ controls the variance of the reverse transition.  
    
    \noindent
    Training is performed by minimizing the simplified noise prediction objective:
    \begin{equation}
        \mathcal{L}_{\text{simple}} = \mathbb{E}_{x_0, \epsilon, t}\Bigl[ \| \epsilon - \epsilon_\theta(x_t, t) \|^2 \Bigr],
    \end{equation}
    which backpropagates gradients through the denoiser to align predicted noise with the ground-truth Gaussian noise added in the forward process.
    
    As shown in Figure~\ref{fig:diffusion-pipeline}, the \emph{upper pathway} corresponds to the forward process, where the clean image is iteratively diffused into noise, while the \emph{lower pathway} illustrates the reverse process, where the U-Net based denoiser progressively reconstructs a clean output from noise. This unified view clarifies how training (noise prediction objective) and inference (reverse sampling) operate within the same probabilistic framework.
    
    \subsubsection{Score-Based and SDE Formulations.}
    An alternative continuous-time view models the diffusion process via stochastic differential equations (SDEs) of the form:
    \begin{equation}
        dx = f(x, t)\,dt + g(t)\,dw,
    \end{equation}
    where $w$ denotes a Wiener process, and $f(x,t)$ and $g(t)$ are drift and diffusion coefficients. The reverse-time SDE can be learned using the time-dependent score function $\nabla_x \log p_t(x)$ estimated through denoising score matching~\cite{song2021scorebased}.
    
    Several LLIE methods adopt this SDE formulation to control information flow and enable entropy-aware transitions during enhancement~\cite{entropy-sde,hou2023global}. These models allow for smoother restoration, better uncertainty modeling, and theoretical connections to energy-based learning and Langevin dynamics. Formulating diffusion as an ordinary differential equation (ODE) or regularized SDE~\cite{song2021scorebased} allows for deterministic sampling and improved control over trajectory dynamics, which has shown benefits in low-light enhancement when stability during denoising is critical.
    
    \subsubsection{Principles and Variants of Diffusion Models}

    To better contextualize diffusion models for LLIE, we formalize key variants commonly observed in recent literature:

    \begin{Summary}[title=\textbf{Vanilla Diffusion (DDPM)}]{}{}
    \begin{definition}
        A diffusion model trained to minimize the denoising objective $\mathcal{L}_{\text{simple}}$ over a fixed variance schedule $\{\beta_t\}$, without explicit conditioning or architectural modifications. Used in works like \textit{DiffLL}~\cite{jiang2023low}, \textit{LightenDiffusion}~\cite{lightendiffusion}, and \textit{DiffLI2D}~\cite{yang2024unleashing}.
    \end{definition}
    \end{Summary}

    \begin{Summary}[title=\textbf{Score-Based Generative Model}]{}{}
    \begin{definition}
        A diffusion model defined by a stochastic process (\textit{e.g.}, Variance Exploding or Variance Preserving SDEs), where generation is performed by solving a reverse SDE using learned score functions. Adopted by models like \textit{Entropy-SDE}~\cite{entropy-sde} and \textit{GSAD}~\cite{hou2023global}.
    \end{definition}
    \end{Summary}

    \begin{Summary}[title=\textbf{Latent Diffusion}]{}{}
    \begin{definition}
        A model where diffusion is applied in a compressed latent space $\mathcal{Z}$ via a learned encoder $E(x)$ and decoder $D(z)$, reducing inference cost:
        \begin{equation}
        \mathcal{L}_{\text{latent}} = \mathbb{E}_{x_0} \left[ \|\epsilon - \epsilon_\theta(E(x_t), t)\|^2 \right].
        \end{equation}
        Used by methods like \textit{L2DM}~\cite{L2DM} and \textit{JoReS-Diff}~\cite{wu2024jores}.
    \end{definition}
    \end{Summary}

    \begin{Summary}[title=\textbf{Conditional Diffusion}]{}{}
    \begin{definition}[Conditional Diffusion]
        A class of diffusion models where the reverse process is guided by external signals (\textit{e.g.}, exposure maps, illumination priors, text prompts):
        \begin{equation}
        p_\theta(x_{t-1} | x_t, c), \quad \text{where } c = \text{guidance input}.
        \end{equation}
        Used in methods like \textit{ExposureDiffusion}~\cite{wang2023exposurediffusion}, \textit{ReDDiT}~\cite{lan2024towards}, and \textit{PSC-Diffusion}~\cite{wan2024psc}.
    \end{definition}
    \end{Summary}

    \subsubsection{Architectural Conditioning for Restoration}

    To adapt diffusion models for image restoration tasks like LLIE, the generative reverse process must be guided by the degraded input (\textit{e.g.}, a low-light image $y$). Several conditioning strategies have been explored in recent literature:

    \begin{itemize}[leftmargin=8pt,itemsep=0.3pt]
        \item \textbf{Input Concatenation:} The low-light image \( y \) is concatenated channel-wise with the noisy sample \( x_t \), allowing the denoising network to attend to both inputs directly.
        \item \textbf{Intermediate Conditioning:} Features extracted from \( y \) are injected into intermediate layers of the denoising network using mechanisms such as cross-attention or feature modulation (\textit{e.g.}, FiLM or AdaIN).
        \item \textbf{Classifier-Free Guidance:} The model is trained both conditionally and unconditionally, allowing for stronger semantic control at inference using a linear combination of conditional and unconditional predictions.
        \item \textbf{Modified Diffusion Initialization:} Some methods initiate the reverse process from a corrupted version of \( y \) instead of pure Gaussian noise, enabling a restoration-centric trajectory.
    \end{itemize}
    
    Most denoising networks \( \epsilon_\theta \) are implemented using U-Net backbones with positional timestep encodings. The U-Net architecture is well suited for LLIE, as its multi-scale design enables the model to capture both global luminance structure and fine-grained textures. Recently, Transformer-based diffusion models (\textit{e.g.}, DiT) have also been explored, offering the ability to capture long-range dependencies that may be useful in complex low-light scenes.
    
    Latent diffusion models further reduce the computational burden by applying diffusion in a compressed latent space. These models rely on an encoder (often a VAE) to map the input to a lower-dimensional representation and a decoder to reconstruct the enhanced image. This design enables high-resolution inference with lower memory and compute requirements while maintaining strong restoration performance.

    \subsection{Theoretical Insights and Broader Foundations}

    \paragraph{Relation to Inverse Problems and Restoration.}
    Low-light image enhancement can be viewed as solving an ill-posed inverse problem, where the observation $y$ (a degraded low-light image) is modeled as the output of a forward degradation process applied to an unknown clean image $x$. That is,
    \begin{equation}
        y = \mathcal{F}(x) + \eta,
    \end{equation}
    where $\mathcal{F}$ is a nonlinear degradation operator (\textit{e.g.}, illumination attenuation, noise injection) and $\eta$ is additive noise. Traditional methods attempt to invert $\mathcal{F}$ using handcrafted priors (\textit{e.g.}, Retinex, contrast stretching), while diffusion models aim to learn the conditional distribution $p(x|y)$ via iterative denoising, offering greater flexibility and robustness.
    
    \paragraph{Variational Interpretation and Objective Functions.}
    The loss function in DDPMs can be interpreted as a variational bound on the negative log-likelihood $\log p_\theta(x_0)$. As shown in~\cite{ho2020ddpm}, the evidence lower bound (ELBO) becomes:
    \begin{equation}
        \mathcal{L}_{\text{ELBO}} = \mathbb{E}_q \left[ \log \frac{q(x_{1:T} | x_0)}{p_\theta(x_{0:T})} \right],
    \end{equation}
    which is decomposed into KL-divergences between the forward and reverse process transitions. In practice, it is simplified to a weighted denoising loss due to tractability.
    
    Alternative formulations include:
    \begin{itemize}[itemsep=0.3pt, left=4pt]
        \item \textbf{Score Matching Loss:} used in score-based SDEs~\cite{song2021scorebased}, optimizing
        \[
            \mathcal{L}_{\text{SM}} = \mathbb{E}_{x \sim p_t} \left[ \|\nabla_x \log p_t(x) - s_\theta(x, t)\|^2 \right].
        \]
        \item \textbf{Likelihood-Based Learning:} applied in models like SFDiff~\cite{wan2024sfdiff}, where log-likelihood of each trajectory is constrained to ensure amplitude fidelity in the Fourier domain.
    \end{itemize}

    \paragraph{Metric Limitations in LLIE.}
    A subtle but important aspect of using generative models like diffusion for LLIE is the inadequacy of traditional metrics. Common metrics such as PSNR and SSIM are known to poorly correlate with perceptual quality~\cite{Zhang2018LPIPS}. In contrast, metrics like LPIPS~\cite{Zhang2018LPIPS} and FID/KID better reflect visual similarity and realism. Several diffusion-based LLIE works (\textit{e.g.}, \textit{DiffLL}~\cite{wavedm}, \textit{LLDiffusion}~\cite{wang2023lldiffusion}) report these perceptual metrics to support their qualitative claims.
    
    \paragraph{Inference Time and Sampling Speed.}
    A recurring challenge in diffusion models is their slow sampling speed due to hundreds of denoising steps. Several works tackle this via:
    \begin{itemize}[label=$\ast$, itemsep=0.3pt, left=4pt]
        \item \textbf{Deterministic Sampling:} DDIM~\cite{song2020denoising} reduces steps via non-stochastic solvers.
        \item \textbf{Latent Diffusion:} reducing dimensionality in latent space (\textit{e.g.}, \textit{JoReS-Diff}~\cite{wu2024jores}, \textit{L2DM}~\cite{L2DM}).
        \item \textbf{Cascaded or Pyramid Samplers:} used in \textit{PyDiff}~\cite{pydiff2023} and \textit{CLE-Diffusion}~\cite{yin2023cle}.
    \end{itemize}
    
    \paragraph{Theoretical Trade-offs in Guidance and Controllability.}
    Conditional diffusion models introduce trade-offs between guidance accuracy and sample diversity. As established in classifier-guided models~\cite{dhariwal2021diffusion}, stronger conditioning improves fidelity but may reduce sample diversity. This trade-off is actively studied in LLIE applications that condition on illumination maps~\cite{lan2024towards}, exposure maps~\cite{wang2023exposurediffusion}, or semantics~\cite{wu2024jores}.
    
    \paragraph{Links to Other Generative Models.}
    Diffusion models differ from GANs and VAEs in key aspects:
    \begin{itemize}[label=$\ast$, itemsep=0.3pt, left=4pt]
        \item \textbf{GANs:} adversarially trained, fast at inference, but suffer from instability and mode collapse.
        \item \textbf{VAEs:} explicit likelihood models, but suffer from blurry reconstructions.
        \item \textbf{Diffusion:} stable training, flexible conditioning, high-quality samples at cost of sampling speed.
    \end{itemize}
    
    Their theoretical convergence to score-matching and energy-based models ties them closely to foundational tools in statistical learning.

%%%%%%%%%%%%%%%%%%%%%%%%%%%%%%%%%%%%%%%%%%%%%%%%%%%%%%%%%%%%%%%%%%%%%%
\section{Taxonomy of Diffusion Models for Low-Light Image Enhancement}
%%%%%%%%%%%%%%%%%%%%%%%%%%%%%%%%%%%%%%%%%%%%%%%%%%%%%%%%%%%%%%%%%%%%%%
\label{sec:taxonomy}

    \begin{figure}
        \centering
        \includegraphics[width=\linewidth]{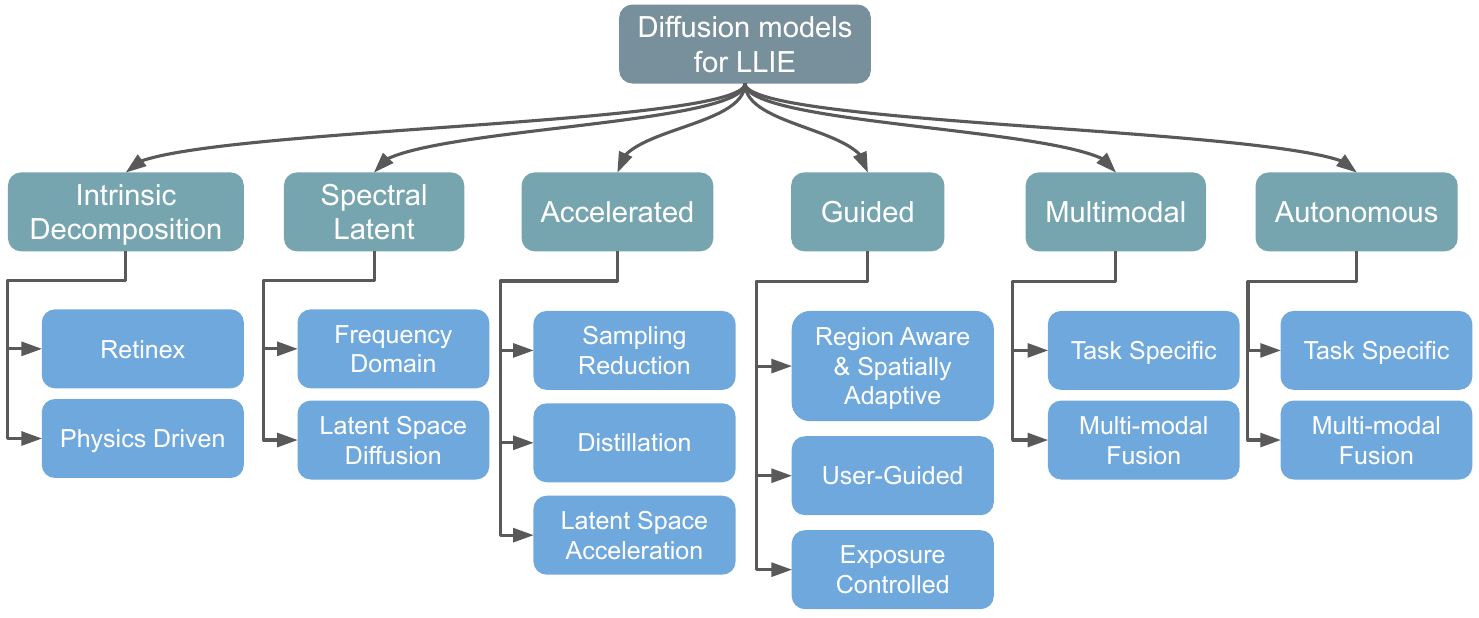}
        \caption{A detailed hierarchical view of the proposed taxonomy for diffusion-based Low-Light Image Enhancement (LLIE) methods. The six primary categories (Intrinsic Decomposition, Spectral \& Latent, Accelerated, Guided, Multimodal, and Autonomous) are shown at the top level. Each category is further broken down into sub-categories representing specific strategic approaches or technical innovations within that domain.}
        \label{fig:flowchart}
    \end{figure}

    Diffusion models (DMs) have significantly advanced low-light image enhancement (LLIE), addressing key challenges such as illumination correction, noise suppression, and detail preservation while balancing computational efficiency. The effectiveness of these models depends on how they integrate domain-specific priors, optimize the reverse process, and adapt to varying enhancement requirements.

    To bring clarity to the diverse methods in this field, we categorize them into six broad groups, each addressing a unique aspect of enhancement while often overlapping with others. 
    This taxonomy, visualized in Figure~\ref{fig:teaser} and Figure~\ref{fig:flowchart}, aims to highlight the primary innovation axis of different approaches.
    Some models focus on intrinsic image decomposition, separating illumination and reflectance for more controlled adjustments. Others operate in spectral or latent domains, transforming images into alternative representations to improve processing efficiency and robustness. Computational cost is a major concern, leading to accelerated diffusion methods that speed up inference through techniques like sampling reduction and distillation. Certain models take a guided approach, incorporating external inputs such as user instructions or semantic prompts for more adaptive enhancements. Meanwhile, modal-specific and multi-modal diffusion models tailor their enhancements for specific tasks, such as text clarity or face restoration, and even integrate alternative sensing modalities like infrared or event cameras. Finally, autonomous diffusion approaches remove the need for paired training data, using self-supervised learning and domain adaptation to generalize across different lighting conditions.

    Our categorization is not rigid, as many models borrow elements from multiple approaches; indeed, a trend of convergence is noticeable where successful techniques from one category are adopted by others. This intermingling suggests a maturation of the field. However, organizing them in this way provides a clearer understanding of the key innovations and focus areas inside diffusion for LLIE methods, thereby helping draw meaningful comparisons between different methodologies. Each category represents a set of design choices aimed at particular aspects of LLIE, but these choices invariably involve trade-offs – a ``\textit{no free lunch}'' principle. For instance, autonomous models achieve data independence but may sacrifice some fidelity compared to supervised counterparts, while guided models offer control but might necessitate more complex conditioning signals.

%%%%%%%%%%%%%%%%%%%%%%%%%%%%%%%%%%%%%%%%%%%%%%%%%%
\subsection{Intrinsic Image Decomposition Methods}
%%%%%%%%%%%%%%%%%%%%%%%%%%%%%%%%%%%%%%%%%%%%%%%%%%

    Intrinsic image decomposition-based diffusion models typically aim to decompose the image components into different sub components (such as \textit{Additive} \& \textit{Multiplicative}, \textit{Illumination} \& \textit{Reflectance}, etc) to more precisely enhance low-light images control those components more strongly.
    These approaches are grounded in theories such as retinex and physics-driven light propagation models, improving interpretability while ensuring stable training. However, their performance depends on accurate prior estimation, making them computationally demanding compared to direct diffusion-based enhancement.

\subsubsection{Retinex Theory Diffusion}
\label{subsec:retinex}

    The Retinex theory models an image $I(x)$ as the product of reflectance $R(x)$ and illumination $L(x)$, i.e., $I(x) = R(x) \cdot L(x)$. This decomposition has been widely used in low-light enhancement, but its direct application often leads to artifacts due to ambiguities in reflectance estimation. Diffusion models offer an alternative by learning the inverse mapping from a degraded image to a well-lit image while incorporating physics-based constraints.

    \begin{figure}
        \centering
        \includegraphics[width=\linewidth]{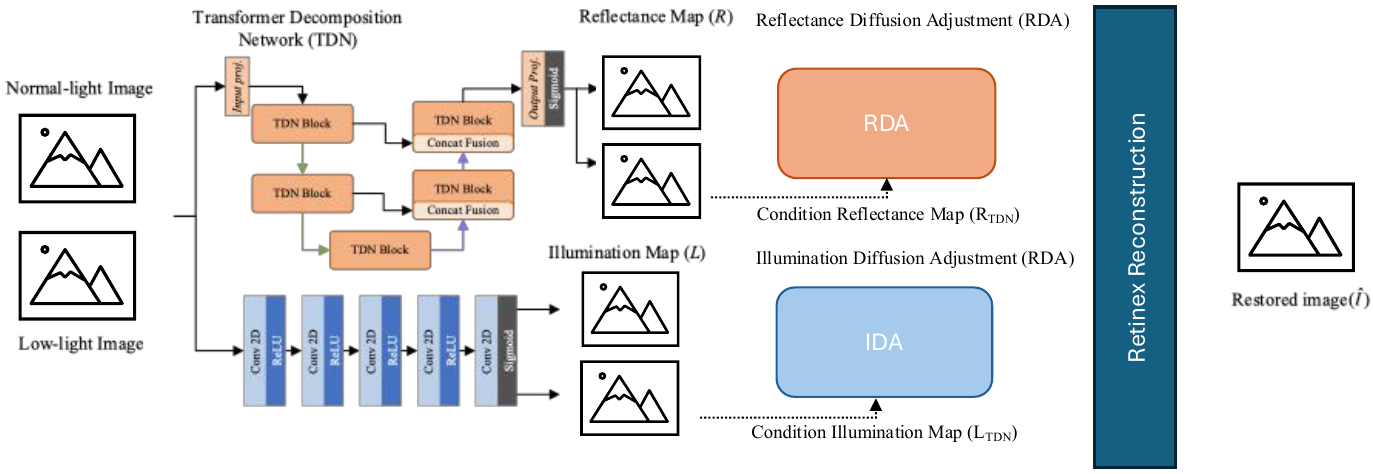}
        \includegraphics[width=\linewidth]{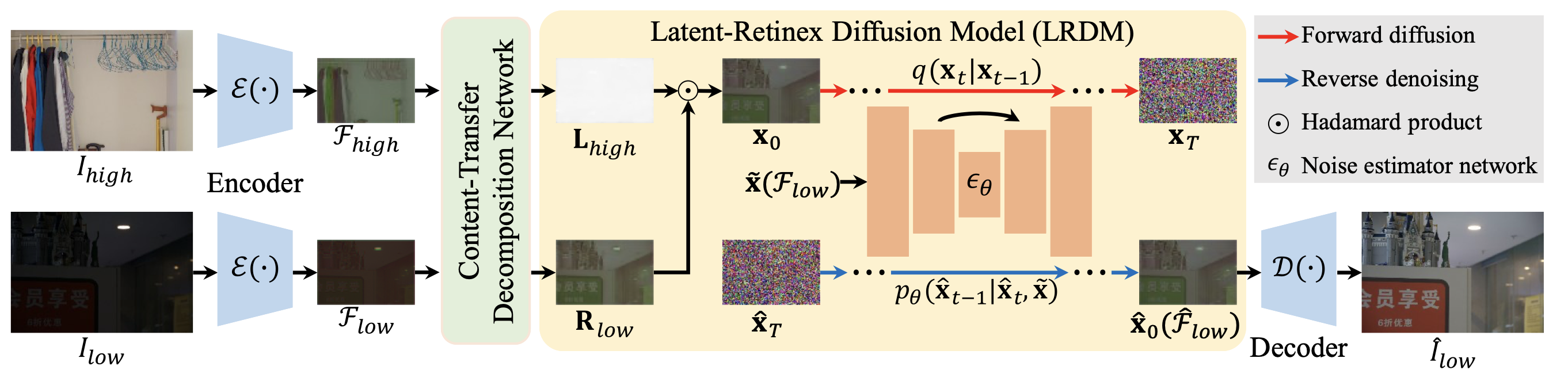}
        \caption{\textbf{[Top]}~The Diff-Retinex paper~\cite{diff-retinex} outlines a modular framework comprising three detachable components: the Transformer Decomposition Network (TDN), Reflectance Diffusion Adjustment (RDA), and Illumination Diffusion Adjustment (IDA). \textbf{[Bottom]}~As presented in the LightenDiffusion~\cite{lightendiffusion} paper, the pipeline uses an encoder $E(\cdot)$ to map unpaired low/normal-light images ($I_{low}$, $I_{high}$) to latent features ($F_{low}$, $F_{high}$), which the CTDN splits into reflectance ($R$) and illumination ($L$) maps. $R_{low}$ and $L_{high}$ drive the forward diffusion, while the reverse denoising process transforms noise $\hat{x}T$ into restored features $\hat{F}{low}$ guided by $F_{low}$ ($\tilde{x}$), then decoded by $D(\cdot)$ into the final image $\hat{I}_{low}$.}
        \label{fig:diff-retinex}
    \end{figure}

	Diff-Retinex~\cite{diff-retinex} introduces retinex decomposition into the diffusion process by conditioning denoising steps on estimated illumination and reflectance maps (as shown in Figure~\ref{fig:diff-retinex}, Top). Instead of treating enhancement as a direct image translation task, the model separately learns to refine illumination and preserve reflectance details. This structured approach prevents over-smoothing and loss of texture, common pitfalls in conventional retinex-based methods.
	EWRD~\cite{ewrd} (Entropy-Weighted Reverse Diffusion) extends this idea by introducing entropy weighting to balance detail preservation and illumination enhancement. It applies a reverse diffusion formulation, refining local pixel statistics based on an entropy constraint that prevents excessive smoothing. The use of entropy as a weighting function allows it to adaptively enhance darker regions without over-amplifying noise. LightenDiffusion~\cite{lightendiffusion} takes a different route by performing Retinex decomposition in a learned latent space rather than directly in the image space (Figure~\ref{fig:diff-retinex}, Bottom). The authors argue that this allows for the generation of content-rich reflectance maps and content-free illumination maps, leading to more effective illumination correction without introducing color distortions. They observe that iteratively refining reflectance maps in this latent space helps reduce artifacts and noise.

    These methods demonstrate that retinex-inspired diffusion models can disentangle illumination from content more effectively than standard diffusion-based enhancement. However, their success depends on accurate decomposition of  $L(x)$ and $R(x)$, which remain a challenging open problem, especially in real-world images characterized by non-uniform lighting, strong cast shadows, or complex material properties where the simple multiplicative model of Retinex theory may not fully hold.

\subsubsection{Physics-Driven Diffusion}
\label{subsec:physic}

    Distinct from, yet related to, Retinex-based methods, physics-driven diffusion models embed explicit physical priors related to light propagation, exposure, and sensor characteristics directly into the diffusion process. These methods attempt to simulate real-world lighting variations and adaptively restore underexposed regions by leveraging photometric consistency constraints or models of light degradation, rather than solely relying on image decomposition.

    \begin{figure}
        \centering
        \includegraphics[width=\linewidth]{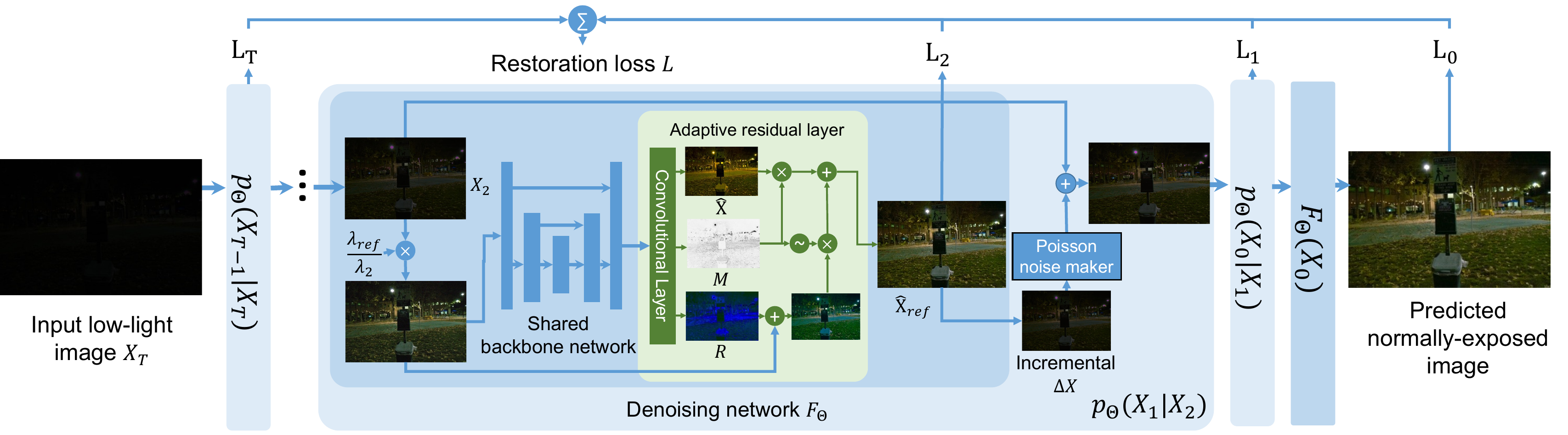}
        \caption{As described in the Exposure Diffusion paper~\cite{wang2023exposurediffusion}, the method uses a progressive refinement framework where a low-light input $X_T$ is gradually enhanced to simulate longer exposures. An adaptive residual layer (green) integrates with any backbone. During both training and inference, intermediate reconstruction losses $L_t$ guide learning, and the final result is $F_{\theta}(X_0)$.}
        \label{fig:exposure-diffusion}
    \end{figure}

    ExposureDiffusion~\cite{wang2023exposurediffusion} conditions the denoising process on exposure-corrected latent variables. Each step of the reverse diffusion process is informed by an estimated exposure adjustment map, allowing for progressive correction of brightness and contrast while aiming to avoid overexposure artifacts, as illustrated in Figure~\ref{fig:exposure-diffusion}. This approach tries to mimic the process of adjusting camera exposure. Zero-LED~\cite{he2024zeroLED} introduces a zero-reference learning paradigm, where the diffusion model learns to enhance images without requiring a well-lit supervision signal. It estimates lighting degradation factors during training and refines the enhancement via frequency-domain reconstruction, leveraging both spatial and spectral cues to ensure perceptual consistency in textures and illumination. Entropy-SDE~\cite{entropy-sde} incorporates a differentiable spatial entropy loss into the diffusion model. This shifts the focus from purely pixel-wise losses towards distribution-based restoration, aiming to improve perceptual quality by encouraging enhanced images to exhibit entropy characteristics similar to those of naturally well-lit images. Such models effectively embed exposure priors, enabling spatially adaptive correction, unlike traditional histogram equalization methods that rely on manually designed constraints, or gamma correction which applies global adjustments.

    % Takeaways
    Such models effectively embed exposure priors, enabling spatially adaptive correction unlike traditional histogram equalization (HE) methods (re-weighting matrix \cite{YingLRWW17}, perceptual importance \cite{WuLHK17}), which rely on manually designed constraints, or gamma correction \cite{severoglu2025acgc} applying global adjustments. However, exposure estimation modules still struggle with complex lighting, such as mixed natural/artificial illumination. LightenDiffusion \cite{lightendiffusion} and Entropy-SDE \cite{entropy-sde} integrate Retinex with diffusion principles, refining illumination correction while mitigating over-enhancement.

    \begin{tcolorbox}[
    enhanced,
    colback=gray!10,
    colframe=gray!40,
    sharp corners,
    boxrule=0.3pt,
    left=4pt,
    right=4pt,
    top=4pt,
    bottom=4pt,
    fonttitle=\bfseries,
    before skip=8pt,
    after skip=8pt,
    boxsep=4pt,
    coltext=black,
    arc=0pt
    ]
    \textbf{Strengths:}
    \begin{itemize}[label=\cmark, itemsep=0.25pt, left=4pt]
        \item Improved \textbf{interpretability} due to \textbf{physics-based priors}.
        \item More \textbf{stable training} compared to purely data-driven approaches.
        \item Lower risk of \textbf{hallucination artifacts} in enhancement.
    \end{itemize}
    \textbf{Limitations:}
    \begin{itemize}[label=\xmark, itemsep=0.25pt, left=4pt]
        \item Requires \textbf{accurate prior estimation} %(Retinex decomposition, exposure maps).
        \item Additional \textbf{computational cost} compared to direct diffusion-based enhancement.
        \item Struggles with scenes where \textbf{retinex assumptions break down} (\textit{e.g.}, heavy specular reflections, extreme non-uniform lighting).
    \end{itemize}
    \end{tcolorbox}

%%%%%%%%%%%%%%%%%%%%%%%%%%%%%%%%%%%%%%%
\subsection{Spectral Latent Diffusion}
\label{subsec:frequency}
%%%%%%%%%%%%%%%%%%%%%%%%%%%%%%%%%%%%%%%

\subsubsection{Frequency-Domain Diffusion}

    Low-light images typically suffer from contrast compression (affecting low-frequency components) and detail loss or noise contamination (affecting high-frequency components). Frequency-domain processing allows models to decompose images into these multi-scale components, enabling targeted enhancement. Unlike pixel-wise operations that might globally affect all image characteristics, frequency-domain methods can selectively adjust different aspects of the image content.

    \begin{figure}
        \centering
        \includegraphics[width=0.47\linewidth]{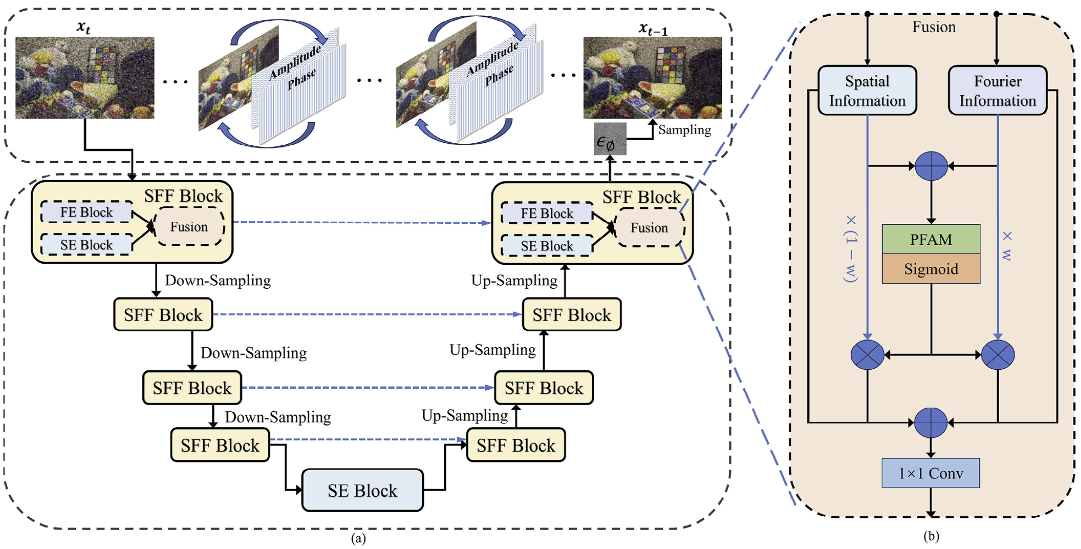}%
        \hspace{0.01\linewidth}%
        \dottedvline{3.8cm}%
        \hspace{0.01\linewidth}%
        \includegraphics[width=0.47\linewidth]{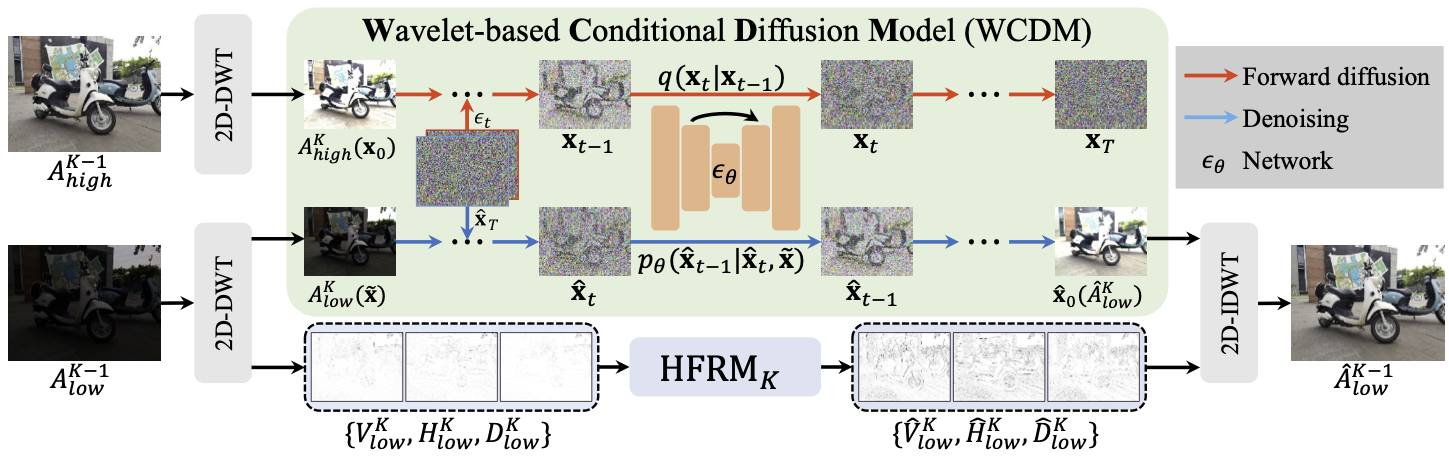}
        \caption{\label{fig:sfdiff}\textbf{[Left]}~In the SFDiff paper~\cite{wan2024sfdiff}, (a) the iterative denoising process applies FFT at both image and feature levels. Spatial and frequency extraction blocks handle respective domains, and Spatial-Frequency Fusion blocks (based on PFAM) assign weights for effective interaction. (b) The fusion block performs weighted blending of spatial and frequency features. \textbf{[Right]}~As described in the DiffLL paper~\cite{jiang2023low}, the pipeline applies 2D-DWT $K$ times to decompose a low-light image into an average coefficient $A^K_{low}$ and $K$ sets of high-frequency components ${V^k_{low}, H^k_{low}, D^k_{low}}$. The proposed WCDM performs diffusion on $A^K_{low}$ for efficient restoration. Forward diffusion is used during training, while denoising occurs in both phases. Final outputs at scale $k-1$ are reconstructed using 2D-IDWT with the restored average and high-frequency coefficients from scale $k$.}
    \end{figure}

\begin{figure}
        \centering
        \includegraphics[width=0.8\linewidth]{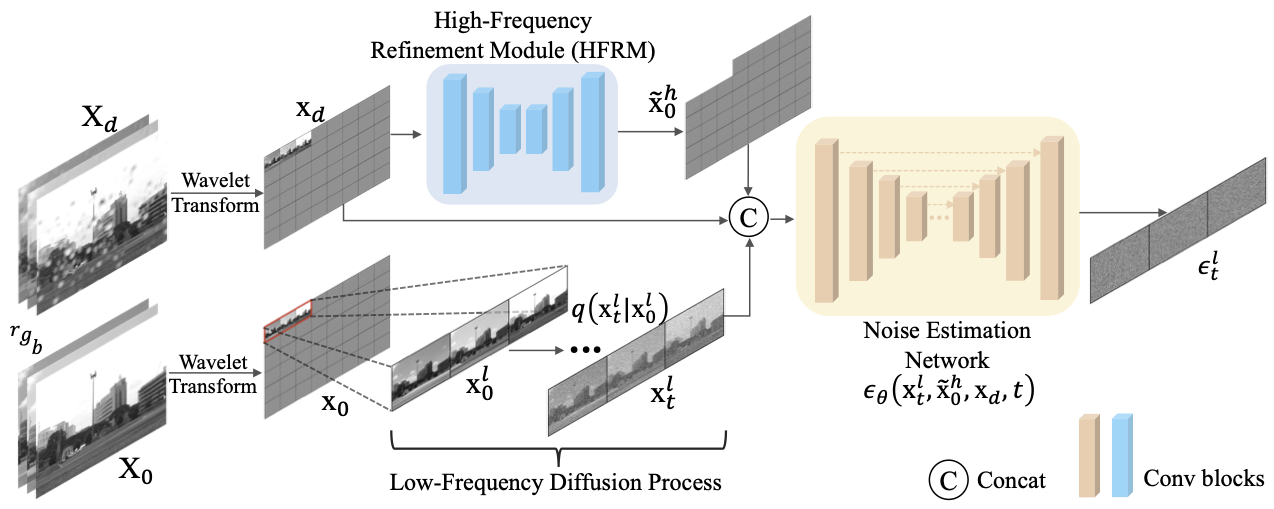}
        \caption{In the WaveDM paper~\cite{wavedm}, degraded and clean RGB images ($X_d$, $X_0$) are transformed into wavelet spectra ($x_d$, $x_0$) via Haar wavelet. The low-frequency spectrum $x^l_0$ produces $x^l_t$ through diffusion, while the high-frequency spectrum $\tilde{x}^h_0$ is estimated from $x_d$ using HFRM. These components, $x_d$, $\tilde{x}^h_0$, and $x^l_t$, are input to $\epsilon_\theta(x^l_t, \tilde{x}^h_0, x_d, t)$ to estimate noise $\epsilon^l_t$ across time steps.}
        \label{fig:wavedm}
    \end{figure}

    SFDiff~\cite{wan2024sfdiff} in Figure~\ref{fig:sfdiff}, incorporates Fourier-domain constraints into the diffusion process, separating amplitude (global brightness) and phase (structural details). The enhancement process refines these components independently, ensuring contrast restoration without structural distortion. By applying constraints in the frequency domain, SFDiff improves global brightness consistency while avoiding texture degradation, a common issue in direct pixel-based diffusion models. Zero-LED~\cite{he2024zeroLED} employs a frequency-guided appearance reconstruction module, leveraging Fourier transforms to estimate degradation patterns in real-world low-light images. Instead of relying on paired training data, it uses unsupervised domain adaptation, learning to bridge the low-light and normal-light frequency distributions. This approach enhances generalization, particularly in unknown lighting conditions where paired data is scarce. DiffLL~(\cite{jiang2023low}, Figure~\ref{fig:sfdiff} right) adopts wavelet-based decomposition, where different frequency bands are processed separately before being recombined. By applying diffusion in multi-scale wavelet space, it achieves better control over detail sharpening and denoising, avoiding over-enhancement in high-exposure regions. WaveDM~\cite{wavedm}, shown in Figure~\ref{fig:wavedm}, uses wavelet-based diffusion modeling, separating low-frequency and high-frequency components to preserve structure and texture. Wavelet transformations allow denoising at different frequency bands, reducing halo artifacts and overexposure. 

    % Takeaways
    These frequency-driven approaches demonstrate that working in transformed domains mitigates the limitations of direct pixel-space diffusion models. By operating on amplitude, phase, or wavelet coefficients, these methods ensure better contrast restoration, noise suppression, and fine-detail preservation. However, they require inverse transformations back to spatial space, which can sometimes introduce artifacts if not well optimized. Methods such as WaveDM~\cite{wavedm} methods enhance efficiency and robustness by working in transformed domains rather than raw pixel space, ensuring better texture preservation and computational efficiency.

%%%%%%%%%%%%%%%%%%%%%%%%%%%%%%%%%%%%%%
\subsubsection{Latent Space Diffusion}
\label{subsec:latent_space}
%%%%%%%%%%%%%%%%%%%%%%%%%%%%%%%%%%%%%%

    Unlike frequency-domain methods that apply explicit transformations, latent-space diffusion models learn a compact, feature-rich representation where enhancement is performed at an abstracted level. This significantly reduces computational overhead, as diffusion steps operate in lower-dimensional latent space rather than full-resolution images.

    \begin{figure}
        \centering
        \includegraphics[width=0.48\linewidth]{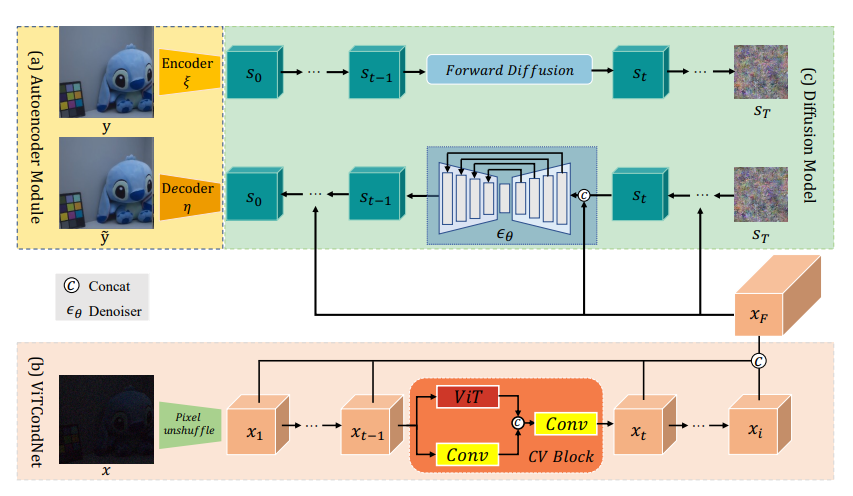}
        \hspace{0.01\linewidth}%
        \dottedvline{3.8cm}%
        \hspace{0.01\linewidth}%
        \includegraphics[width=0.48\linewidth]{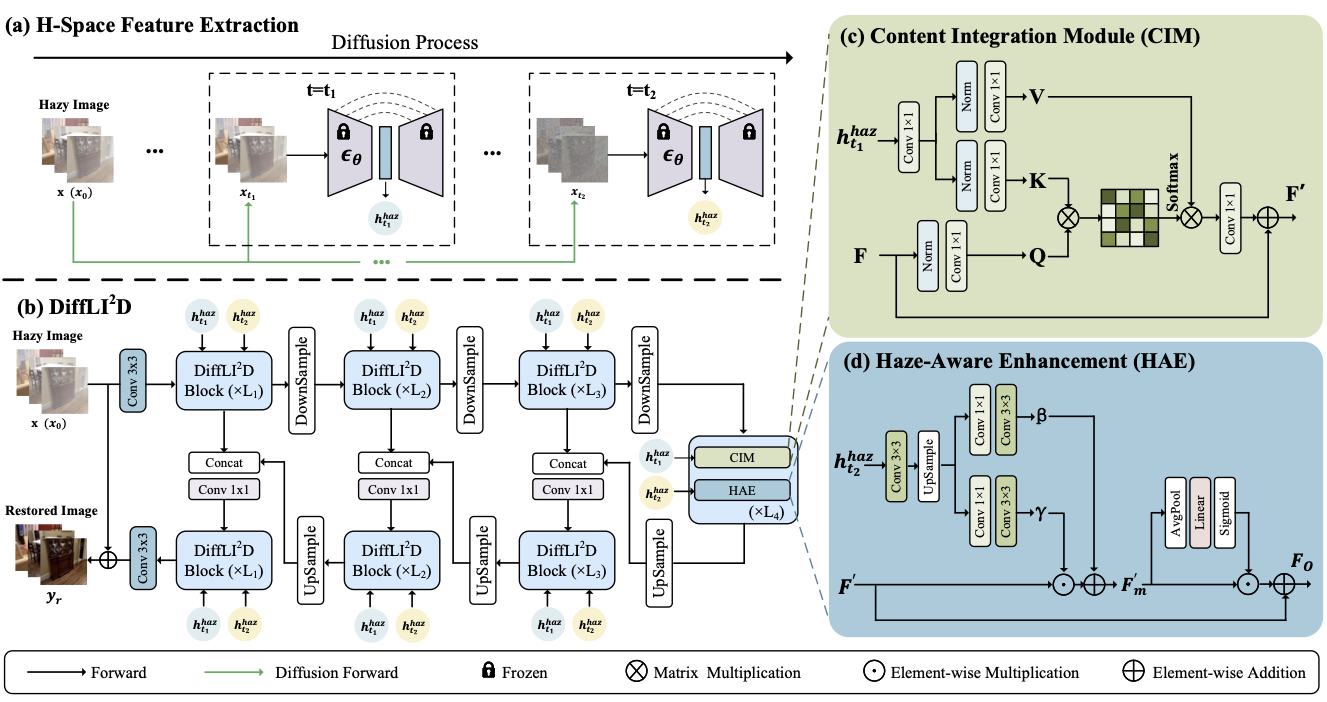}
        \caption{\textbf{[Left]}~In the L2DM paper~\cite{L2DM}, (a) a trainable autoencoder maps images from pixel to latent space, reducing computation. (b) The proposed condition module offers efficient feature extraction with fewer parameters. (c) The condition feature map is concatenated with $s_t$ and input to the time-conditional U-Net. \textbf{[Right]}~In the DiffLI$^2$D paper~\cite{yang2024unleashing}, (a) a hazy image $x$ is perturbed into $x_{t1}$ and $x_{t2}$ (Eq.~\ref{eq:forward}), then passed through a frozen diffusion model to obtain h-space features $h^{haz}{t1}$ and $h^{haz}{t2}$. (b) The U-Net-based dehazing network includes Content Integration (CIM) and Haze-Aware Enhancement (HAE) modules, using $h^{haz}{t1}$ for content recovery and $h^{haz}{t2}$ for haze removal.}
        \label{fig:L2DM-DiffLI2D}
    \end{figure}

    L2DM~\cite{L2DM} (also discussed in \ref{subsec:latent_space}) applies diffusion in a learned latent space, rather than pixel space, dramatically reducing memory and computational costs. By training an autoencoder that learns an optimal low-light representation, L2DM performs enhancement directly on compressed feature maps, leading to faster inference. This method is particularly useful for real-time applications where standard diffusion models are prohibitively slow. As seen on the right side of Figure~\ref{fig:L2DM-DiffLI2D}, DiffLI$^2$D~\cite{yang2024unleashing} leverages pre-trained frozen diffusion models and extracts features from their semantic latent space. Instead of training a diffusion model from scratch, it reuses existing diffusion networks and learns how to navigate their latent space to restore degraded images. This reduces the need for extensive re-training, making it a plug-and-play solution for enhancement tasks.

    % Takeaways
    Latent-space diffusion significantly improves computational efficiency while maintaining strong restoration quality. However, the challenge lies in maintaining spatial consistency, since enhancement occurs in compressed representations, reconstructing fine details in the final image can be difficult. Additionally, these models require a well-trained encoder-decoder to ensure that low-light degradations are accurately encoded in latent space. Such methods enhance efficiency and robustness by working in transformed domains rather than raw pixel space, ensuring better texture preservation and computational efficiency.

    \begin{tcolorbox}[colback=gray!10, colframe=gray!40, sharp corners, boxrule=0.3pt, left=2pt, right=2pt, top=2pt, bottom=2pt]
    \textbf{Strengths:}
    \begin{itemize}[label=\cmark, itemsep=0.25pt, left=4pt]
        \item \textbf{Frequency-domain methods} provide \textbf{structural consistency} and avoid over-enhancement.
        \item \textbf{Latent-space diffusion} is \textbf{computationally efficient}, making it scalable for real-time applications.
        \item Both approaches improve \textbf{robustness} by working in spaces where \textbf{low-light distortions are more clearly defined}.
    \end{itemize}

    \textbf{Limitations:}
    \begin{itemize}[label=\xmark, itemsep=0.25pt, left=4pt]
        \item Frequency-based models require \textbf{careful inverse transformations}, as spectral reconstruction can introduce \textbf{halo artifacts}.
        \item Latent-space diffusion models depend on \textbf{encoder quality}, and poor latent representations can lead to \textbf{loss of fine details}.
        \item Some spectral methods (\textit{e.g.}, Fourier-based approaches) may fail in images with \textbf{extreme non-uniform lighting}, as frequency decomposition assumes \textbf{global consistency} in degradation patterns.
    \end{itemize}
    \end{tcolorbox}

%%%%%%%%%%%%%%%%%%%%%%%%%%%%%%%%%%
\subsection{Accelerated Diffusion}
\label{subsec:efficiency}
%%%%%%%%%%%%%%%%%%%%%%%%%%%%%%%%%%

    Diffusion models offer superior image enhancement quality but suffer from high computational costs due to their iterative denoising process \cite{ho2020ddpm}. Standard diffusion-based low-light enhancement requires hundreds to thousands of reverse steps, making real-time deployment impractical. To overcome this, accelerated diffusion methods focus on reducing sampling steps, optimizing the reverse process, and leveraging efficient latent-space representations. These approaches aim to maintain enhancement quality while significantly lowering inference time, making diffusion models viable for applications in edge computing, mobile devices, and real-time video processing.

\subsubsection{Sampling Reduction and Trajectory Optimization}

    Reducing the number of diffusion steps is critical for making enhancement models real-time. Many sampling acceleration techniques exist, but their direct application often leads to image artifacts and degraded enhancement quality \cite{lan2024towards,pydiff2023,L2DM}. To address this, optimized trajectory refinement techniques aim to preserve visual fidelity while minimizing the number of required sampling iterations.

    ReDDiT~\cite{lan2024towards} (Reflectance-Aware Diffusion with Distilled Trajectory) introduces reflectance-aware trajectory refinement (RATR) to address the performance degradation caused by fewer sampling steps. Instead of directly reducing the steps, it refines the noise trajectory to align it with reflectance-based priors, improving detail retention even with fewer iterations. The method achieves SOTA performance using as few as 2 steps, compared to standard diffusion models that require hundreds. $L^{2}DM$~\cite{L2DM} employs latent-space diffusion not only for efficiency but also to optimize trajectory-based sampling. By operating in a compressed feature space, it minimizes the number of required diffusion iterations while maintaining structural coherence. DiffLLE~\cite{yang2023difflle}, shown on the left of Figure~\ref{fig:difflle-pydiff}, introduced diffusion-guided domain calibration (DDC) to adapt to real-world illumination inconsistencies, ensuring that models do not overcorrect or under-enhance regions based on unrealistic training distributions.

    % Takeaways
    These methods show that naively reducing sampling steps degrades performance, but trajectory refinement and latent-space processing help mitigate quality loss. However, designing effective noise trajectory optimizations remains a challenge, as aggressive step reduction can still lead to blurring and loss of fine details.

\subsubsection{Distillation and Knowledge Transfer for Faster Inference}

    Distillation-based approaches train a student model to mimic the behavior of a full-fledged diffusion model but with significantly fewer denoising steps. Unlike standard acceleration methods, which focus on trajectory refinement, these methods compress knowledge into a lightweight model that can perform enhancement in fewer iterations.

    \begin{figure}
        \centering
        \includegraphics[height=0.35\linewidth]{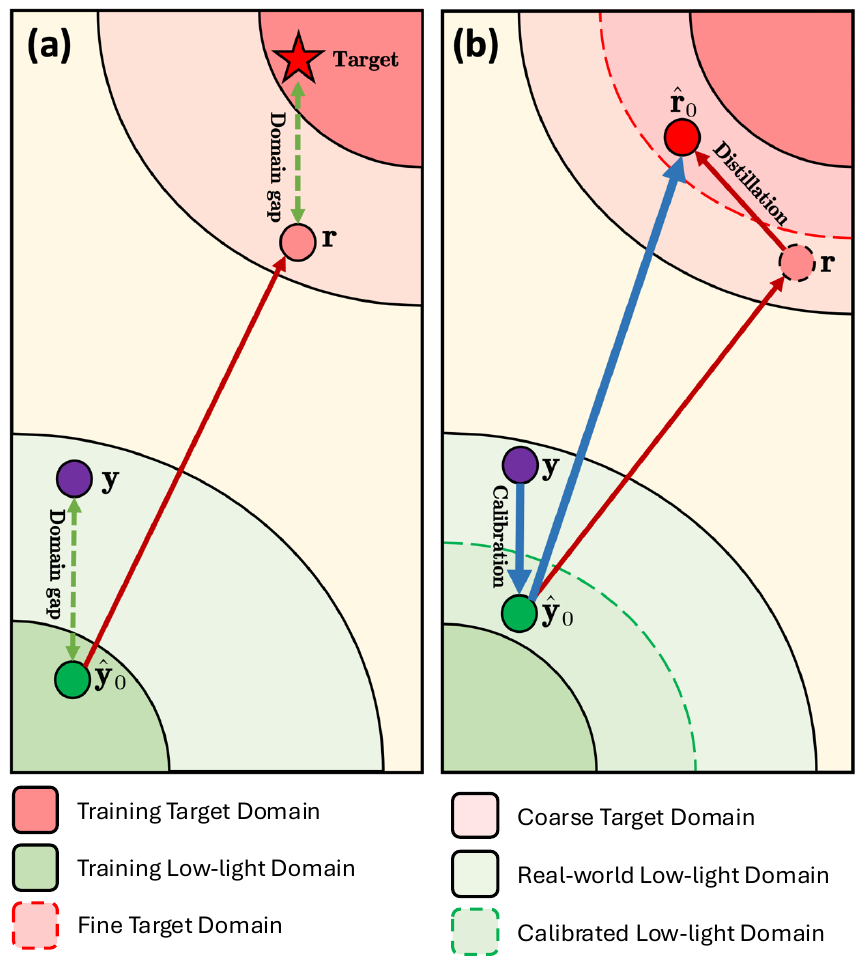}
        \hspace{0.01\linewidth}%
        \dottedvline{5.8cm}%
        \hspace{0.01\linewidth}%
        \includegraphics[height=0.36\linewidth]{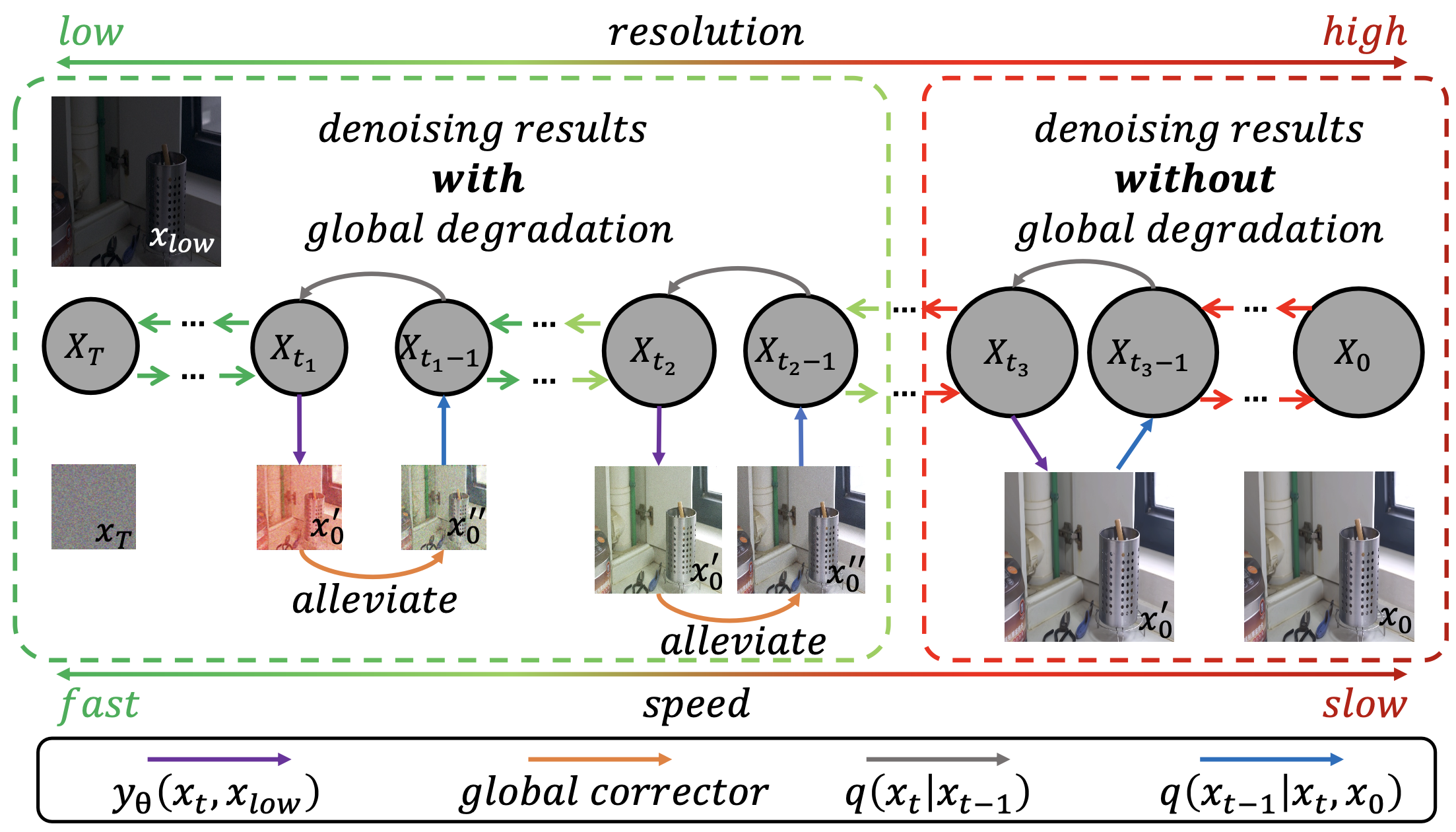}
        \caption{\label{fig:difflle-pydiff}\textbf{[Left]}~ Domain representation and enhancement workflow of DiffLLE~\cite{yang2023difflle} \textbf{(a)} Conventional methods enhance training low-light inputs to coarse normal-light domains, but fail to bridge the domain gap to the true enhancement target. \textbf{(b)} DiffLLE introduces domain calibration and distillation, aligning real-world low-light inputs with the training domain, then guiding them to the fine target domain. The green, blue, and red arrows represent calibration, inference, and distillation paths, respectively. \textbf{[Right]}~Overview of the proposed PyDiff framework~\cite{pydiff2023}. The denoising network $y_\theta(x_t, x_{\text{low}}$) estimates an approximation of $x_0$, as described in Equation-\ref{eq:eq8-pydiff}. For visualization clarity, the input $x_{\text{low}}$ has been brightened.}
    \end{figure}

    \begin{equation}\label{eq:eq8-pydiff}
        y_{\theta}(x_t, x_{\text{low}}) = \frac{1}{\sqrt{\bar{\alpha}_t}} \left( x_t - \sqrt{1 - \bar{\alpha}_t} \cdot \epsilon_{\theta}(x_t, x_{\text{low}}, t) \right)
    \end{equation}

    PyDiff~\cite{pydiff2023}, as seen on the right side of Figure~\ref{fig:difflle-pydiff} applies pyramid-based knowledge distillation, where the model learns to enhance low-light images across multiple resolution scales. This progressive refinement strategy allows enhancement to be computed in lower resolutions first, before being upsampled to finer details, reducing the overall computational burden. Specifically, when low-light images translate to normal light by translating $z$ to $y$ as shown in Eq.~\ref{eq:eq8-pydiff}. MDMS~\cite{mdms} proposed a Multi-Domain Multi-Scale (MDMS) strategy, integrating spatial and frequency-domain features while using multi-resolution sampling to eliminate checkerboard artifacts common in single-scale patch-based diffusion~\cite{pydiff2023} and transformer~\cite{wang2023ultra} models. ReDDiT~\cite{lan2024towards} also integrates a distillation framework, where a compact model is trained to approximate the full diffusion trajectory using a small number of optimized steps. This reduces the number of function evaluations required per sample while preserving high-fidelity reflectance reconstruction. 

    % Takeaways
    These methods demonstrate that distilling diffusion knowledge into smaller models or multi-resolution strategies enables faster inference without fully sacrificing quality. However, distillation may not fully capture the generative diversity of full diffusion models, leading to limited flexibility in extreme low-light conditions.

\subsubsection{Latent-Space Acceleration for Real-Time Enhancement}

    Some approaches focus on optimizing the underlying representation space to improve efficiency rather than modifying the diffusion process itself. These methods operate in compressed latent spaces, reducing the dimensionality of inputs and allowing denoising steps to be computed in a lower-dimensional domain.

	$L^{2}DM$~\cite{L2DM} (also discussed in \ref{subsec:latent_space}) achieves speedup by mapping low-light images to a compact feature space where diffusion operates more efficiently. This reduces both memory footprint and computational cost, making diffusion-based enhancement feasible for real-time applications.
	DiffLI$^2$D~\cite{yang2024unleashing} uses a pre-trained latent-space representation from existing diffusion models to bypass redundant training. By learning how to navigate the latent space of a pre-trained model, it eliminates the need for extensive retraining, allowing faster adaptation to new lighting conditions.

    % Takeaways 
    Latent-space acceleration significantly improves efficiency while maintaining enhancement quality, but it depends heavily on encoder performance; poorly trained encoders may fail to retain fine details, leading to loss of textural fidelity in enhanced images.

    \begin{tcolorbox}[colback=gray!10, colframe=gray!40, sharp corners, boxrule=0.3pt, left=2pt, right=2pt, top=2pt, bottom=2pt]
    \textbf{Strengths:}
    \begin{itemize}[label=\cmark, itemsep=0.25pt, left=4pt]
        \item Reduces inference time, making real-time enhancement feasible.
        \item Trajectory refinement preserves reflectance structure in low-light images.
        \item Latent-space acceleration reduces computational overhead while maintaining enhancement quality.
    \end{itemize}

    \textbf{Limitations:}
    \begin{itemize}[label=\xmark, itemsep=0.25pt, left=4pt]
        \item Aggressive step reduction can lead to blurry outputs and detail loss.
        \item Distilled models may struggle with generalization across unseen lighting conditions.
        \item Latent-space methods require high-quality encoders to retain structural integrity.
    \end{itemize}
    \end{tcolorbox}

%%%%%%%%%%%%%%%%%%%%%%%%%%%%%
\subsection{Guided Diffusion}
\label{subsec:adaptive}
%%%%%%%%%%%%%%%%%%%%%%%%%%%%%

    Low-light enhancement is often context-dependent, meaning different parts of an image may require varying levels of correction based on scene content, user intent, or task-specific constraints. Standard diffusion models apply uniform enhancement, leading to over-exposed regions, excessive smoothing, or loss of desired ambient lighting. Guided diffusion models address this by incorporating explicit control mechanisms, enabling spatially adaptive, user-interactive, or semantically driven enhancements. These approaches improve flexibility and interpretability, making diffusion-based enhancement more practical for real-world applications.

\subsubsection{Region-Aware and Spatially Adaptive Diffusion}

    Different regions in an image experience different levels of degradation, necessitating local enhancements rather than a uniform correction across the entire image. Region-aware guided diffusion models allow spatially adaptive transformations, ensuring that well-lit areas remain untouched while darker regions receive targeted improvements.
	CLE-Diffusion~\cite{yin2023cle} introduces controllable local enhancement by conditioning diffusion on region-based masks generated using Segment Anything Model (SAM)~\cite{kirillov2023segment}. This ensures that localized low-light regions receive enhancement without globally altering the entire image. Fused segmentation maps into the diffusion process allow the enhancement to be spatially constrained, thereby avoiding overexposure or loss of fine details in well-lit areas.
	ExposureDiffusion~\cite{wang2023exposurediffusion} uses local exposure priors to guide the enhancement, making adaptive corrections in underexposed regions while preserving naturally well-lit portions. By leveraging scene-specific exposure cues, it ensures contrast restoration without unnatural lighting shifts.

    % Takeaways
    These methods demonstrate that region-based conditioning prevents excessive global transformations, making low-light enhancement more controlled and natural-looking. However, their effectiveness depends on accurate segmentation and exposure estimation, and if the masks are inaccurate, the enhancement may be misaligned with actual lighting conditions.

\begin{figure}[ht!]
   \centering
   \includegraphics[width=\linewidth]{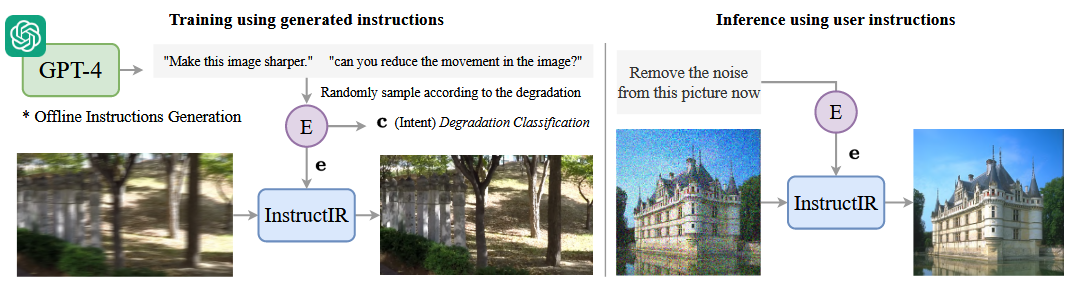}
   \includegraphics[width=\linewidth]{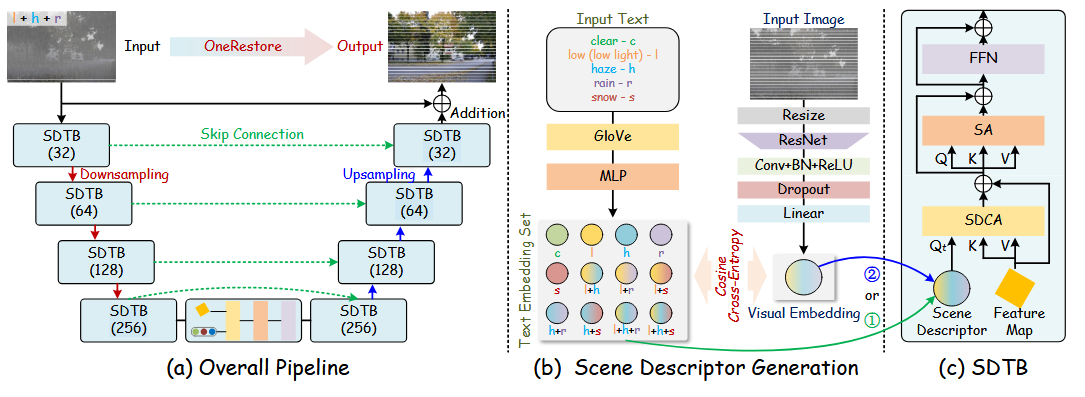}
   \caption{\textbf{[Top]}~The InstructIR framework~\cite{conde2024instructir} demonstrates instruction-based enhancement through a two-phase approach: during training, GPT-4 generates diverse degradation-specific prompts for self-supervised learning, while at inference time, users provide natural language instructions (e.g., ``Remove the noise from this picture now'') to guide the restoration process. The encoder $E$ maps degraded images to embeddings $e$, which InstructIR processes according to the provided textual guidance. \textbf{[Bottom]}~The OneRestore architecture~\cite{guo2024onerestore} presents a unified multi-task restoration pipeline where (a) the overall U-Net structure with SDTB blocks processes inputs through skip connections at resolutions 32, 64, 128, and 256 channels, (b) scene descriptor generation combines text embeddings with visual features through GloVe and MLP layers to produce contextual guidance, and (c) the Scene Descriptor-guided Transformer Block (SDTB) integrates spatial self-attention (SA) and cross-attention mechanisms with scene descriptors to enable task-specific enhancement conditioning.}
   \label{fig:instructir-onerestore}
\end{figure}

\subsubsection{User-Guided and Instruction-Based Diffusion}

    While physics-based guidance improves adaptability, some scenarios require explicit human control over enhancement. Instruction-based diffusion models incorporate natural language prompts or user-defined constraints to customize enhancement outputs.
    
    InstructIR~\cite{conde2024instructir}, displayed in the top portion of Figure~\ref{fig:instructir-onerestore}, brings vision-language models (VLMs) into the enhancement pipeline, enabling users to provide text-based instructions (\textit{e.g.}, ``brighten shadows,'' ``reduce noise in dark areas'') to guide the enhancement process. By conditioning the denoising process on semantic descriptions, it allows for intuitive and interactive control, making it suitable for content-specific restoration.

    OneRestore~\cite{guo2024onerestore}, illustrated in the bottom of Figure~\ref{fig:instructir-onerestore}, extends user-guided control by introducing multi-task conditioning, where the enhancement is modulated based on scene type (\textit{e.g.}, ``urban night scene'' vs. ``indoor low-light''). This approach ensures that enhancements align with contextual needs, rather than applying a one-size-fits-all correction.
    
    % Takeaways
    Instruction-based guidance introduces semantic awareness into low-light enhancement, but its success depends on robust prompt understanding. In cases where scene semantics are ambiguous, text-based models may misinterpret enhancement requirements, leading to suboptimal restorations.

\subsubsection{Exposure and Lighting-Controlled Diffusion}

    Controlling the level of enhancement is crucial in practical applications, where over-enhancement can lead to unnatural results. Exposure-aware diffusion models allow adjustment of brightness levels using explicit exposure control mechanisms.

    ExposureDiffusion~\cite{wang2023exposurediffusion} (also mentioned in Region-Aware Diffusion) models the exposure distribution of real-world images and conditions the denoising process on exposure priors. This ensures that enhancement follows natural lighting patterns, avoiding excessive brightening that distorts scene realism.

    Zero-LED~\cite{he2024zeroLED} (discussed in Spectral Latent Diffusion) introduces a zero-reference lighting estimation module, allowing diffusion models to infer optimal exposure correction without paired training data. By learning from statistical distributions of normal-lit images, it applies enhancement in an adaptive and reference-free manner.

    % Takeaways
    Exposure-guided diffusion models ensure natural-looking enhancements, but they require accurate exposure estimation. Errors in exposure modeling may cause inconsistencies in lighting correction, particularly in highly non-uniform illumination environments.

    \begin{tcolorbox}[colback=gray!10, colframe=gray!40, sharp corners, boxrule=0.3pt, left=2pt, right=2pt, top=2pt, bottom=2pt]
    \textbf{Strengths:}
    \begin{itemize}[label=\cmark, itemsep=0.25pt, left=4pt]
        \item Improves spatial adaptability, preventing over-enhancement in well-lit regions.
        \item Enables user interaction, making enhancement customizable.
        \item Exposure-aware conditioning ensures more natural brightness adjustments.
    \end{itemize}

    \textbf{Limitations:}
    \begin{itemize}[label=\xmark, itemsep=0.25pt, left=4pt]
        \item Requires accurate segmentation or prompts, which may introduce errors.
        \item Instruction-based methods struggle with ambiguous or poorly phrased commands.
        \item Exposure models depend on robust lighting estimation, which may fail in extreme conditions.
    \end{itemize}
    \end{tcolorbox}

%%%%%%%%%%%%%%%%%%%%%%%%%%%%%%%%%%%%%%%%%%%%%%%%%%%%%
\subsection{Modal-Specific and Multi-Modal Diffusion}
\label{subsec:specialized}
%%%%%%%%%%%%%%%%%%%%%%%%%%%%%%%%%%%%%%%%%%%%%%%%%%%%%

\subsubsection{Diffusion for Task-Specific Low-Light Enhancement}

    Low-light image enhancement is typically addressed using single-modal inputs, standard RGB images. However, this ignores additional information available from different imaging sensors or task-specific constraints. Modal-specific diffusion models specialize in enhancing low-light images for particular downstream tasks (\textit{e.g.}, OCR~\cite{lin2025text}, object detection~\cite{xu2021exploring,xu2025upt}, semantic segmentation~\cite{yao2024event,xu2025upt}, face detection~\cite{yang2025mdff}), while multi-modal diffusion models incorporate alternative sensing modalities such as event cameras, infrared (IR), or depth data to compensate for missing information in extreme lighting conditions. These approaches broaden the applicability of diffusion models beyond standard visual restoration, making them more robust in real-world scenarios.

    Standard enhancement models prioritize visual fidelity, but for task-driven applications such as scene text recognition, object detection, and segmentation, a different optimization objective is required, one that preserves critical task-relevant features rather than focusing purely on perceptual quality.

\begin{figure}[ht!]
    \centering
    \includegraphics[width=\linewidth]{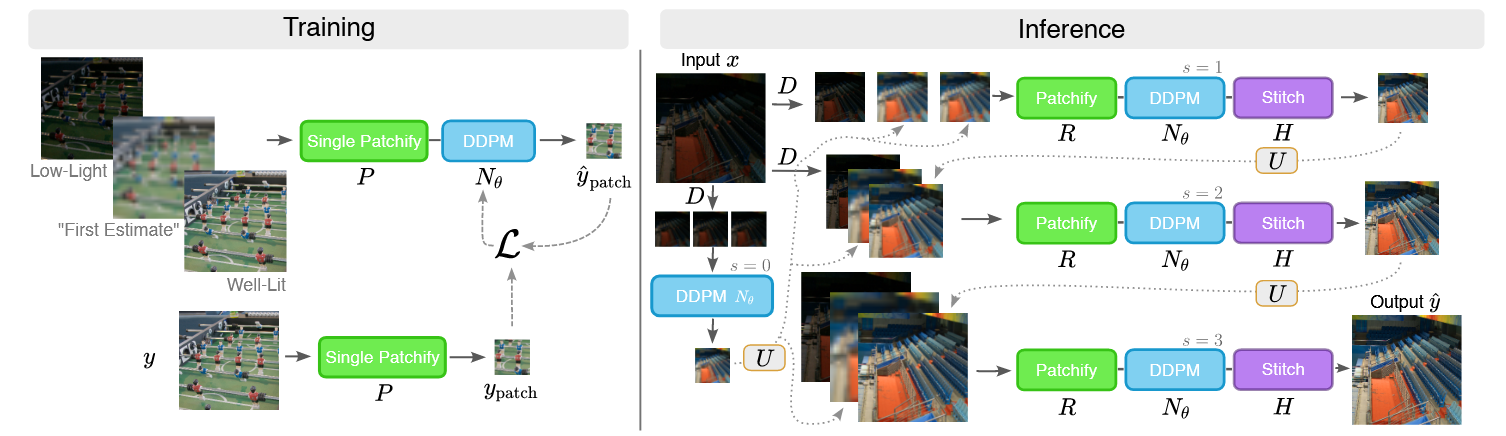}
    \includegraphics[width=\linewidth]{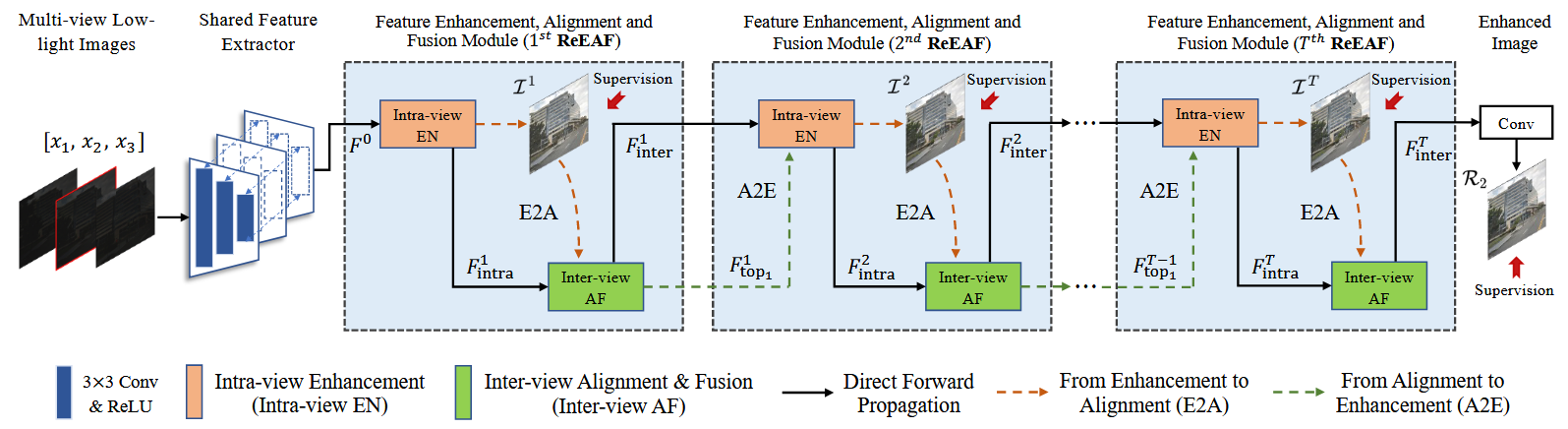}
    \caption{\textbf{[Top]}~The DID pipeline~\cite{nguyen2024diffusion} employs a multi-scale patch-based approach where training involves randomly cropped $32 \times 32$ patches at multiple scales (denoted by $s$) using Single Patchify. Low-light patches, low-resolution well-lit patches, and high-resolution well-lit patches are concatenated as conditioning inputs for the DDPM network $N_{\theta}$ to denoise and reconstruct well-lit patches $\hat{y}_{patch}$. During inference, the trained DDPM operates on 4 scales successively, with each prediction upsampled using $U$ and refined through Patchify, DDPM, and Stitch operations to achieve progressively enhanced resolution in the final output $\hat{y}$. \textbf{[Bottom]}~The RCNet framework~\cite{luo2025rcnet} processes multi-view low-light images $[x_1, x_2, x_3]$ through a Shared Feature Extractor, followed by $T$ recurrent Feature Enhancement, Alignment and Fusion (ReEAF) modules. Each ReEAF module contains Intra-view Enhancement (Intra-view EN) and Inter-view Alignment \& Fusion (Inter-view AF) components that progressively refine features through direct forward propagation, Enhancement-to-Alignment (E2A), and Alignment-to-Enhancement (A2E) pathways, ultimately producing an enhanced single image $R_2$ from the primary view $x_2$ via supervision signals at each recurrent stage.}
    \label{fig:multimodal-diffusion}
\end{figure}

    Diffusion-in-the-Dark~\cite{nguyen2024diffusion}, shown in the top of Figure~\ref{fig:multimodal-diffusion}, targets text readability enhancement in low-light images. Unlike generic enhancement models that optimize for human perception, this model reconstructs low-light text with explicit constraints on stroke clarity, edge preservation, and OCR consistency. The model’s denoising process incorporates text-structure priors, improving downstream text recognition performance compared to standard image restoration models. Zero-LED~\cite{he2024zeroLED} extends the idea of zero-reference enhancement to task-aware restoration, leveraging frequency-domain constraints to enhance both global contrast and character-level structure, making it effective for OCR and small-object recognition tasks. OneRestore~\cite{guo2024onerestore} explores low-light enhancement in the presence of multiple degradations (\textit{e.g.}, fog, rain, and low visibility). By integrating task-specific priors, the model learns to enhance images in adverse conditions while preserving object-level integrity, making it useful for autonomous driving and surveillance.

    % Takeaways:
    These models demonstrate that enhancement should be conditioned on the end task rather than purely on visual realism. However, their effectiveness depends on precisely defining task constraints, for instance, enhancing text may not require the same processing as improving face detection in low-light conditions.

\subsubsection{Multi-Modal Fusion for Robust Low-Light Enhancement}
\label{subsec:multi-modal}

    In extreme low-light scenarios, RGB cameras struggle due to photon noise, motion blur, and low dynamic range. To overcome these limitations, multi-modal diffusion models integrate alternative sensing modalities, such as event cameras, infrared (IR), and depth sensors, to recover missing details that traditional cameras fail to capture.

    EVSNet~\cite{yao2024event} fuses event-based vision with conventional RGB imaging to improve low-light video segmentation. Unlike frame-based cameras, event cameras detect brightness changes at a high temporal resolution, making them resilient to motion blur and extreme lighting. The model integrates event-based motion cues into the diffusion process, refining temporal coherence and structure preservation for low-light video applications. RCNet~\cite{luo2025rcnet}, illustrated in the bottom portion of Figure~\ref{fig:multimodal-diffusion}, addresses multi-view low-light enhancement, where multiple perspectives of the same scene are combined to infer missing details. Instead of treating images independently, this model learns to fuse complementary information across different viewpoints, enhancing object clarity and depth consistency. AutoDir~\cite{jiang2024autodir} explores self-supervised enhancement across multiple imaging modalities, dynamically selecting the best modality (RGB, IR, or depth) to guide the diffusion-based restoration process. This makes it highly adaptable for night-time surveillance and thermal imaging applications. AGLLDiff~\cite{lin2024aglldiff} proposed a training-free unsupervised framework, guiding diffusion models via attribute constraints (exposure, structure, color) instead of relying on predefined degradation models. EL2NM~\cite{qin2024el2nm} uses extreme low-light noise distributions using cold diffusion iteration for adaptive noise handling and denoising under near-zero-light conditions.

    % Takeaways 
    Multi-modal fusion significantly enhances robustness under extreme lighting conditions, but it comes with a higher computational cost due to the need for cross-modal alignment. Additionally, multi-modal datasets are limited, making supervised training challenging.

    \begin{tcolorbox}[colback=gray!10, colframe=gray!40, sharp corners, boxrule=0.3pt, left=2pt, right=2pt, top=2pt, bottom=2pt]
    \textbf{Strengths:}
    \begin{itemize}[label=\cmark, itemsep=0.25pt, left=4pt]
        \item Improves low-light enhancement for specific tasks such as OCR, object detection, and segmentation.
        \item Multi-modal approaches compensate for RGB limitations in extreme lighting conditions.
        \item Event-based fusion improves video consistency in low-light settings.
    \end{itemize}

	\textbf{Limitations:}
    \begin{itemize}[label=\xmark, itemsep=0.25pt, left=4pt]
        \item Higher computational complexity due to multi-modal fusion.
        \item Limited availability of multi-modal training datasets hinders scalability.
        \item Task-specific models lack generalization, requiring custom retraining for new applications.
    \end{itemize}
    \end{tcolorbox}

%%%%%%%%%%%%%%%%%%%%%%%%%%%%%%%%%
\subsection{Autonomous Diffusion}
\label{subsec:autonomous}
%%%%%%%%%%%%%%%%%%%%%%%%%%%%%%%%%

    Traditional low-light enhancement models rely on paired training datasets, where a degraded low-light image is mapped to its well-lit counterpart. However, real-world low-light conditions exhibit unpredictable degradations that are often scene-dependent, sensor-specific, and non-uniform. Collecting high-quality paired data is challenging, making fully supervised approaches impractical.

    Autonomous diffusion models address this by leveraging self-supervised learning, zero-reference adaptation, and unsupervised domain alignment to enhance low-light images without requiring paired training data. These models offer higher generalization, adaptability to unseen lighting conditions, and independence from supervised constraints, making them more scalable and robust in diverse real-world scenarios.

\subsubsection{Zero-Shot and Self-Supervised Learning in Diffusion Models}

    Traditional supervised learning assumes that low-light degradations follow a predictable pattern, but in reality, they vary significantly across cameras, environments, and exposure settings. Zero-shot and self-supervised methods learn to enhance low-light images without explicitly relying on ground-truth well-lit images.
	
\begin{figure}[ht!]
   \centering
   \includegraphics[width=.9\linewidth]{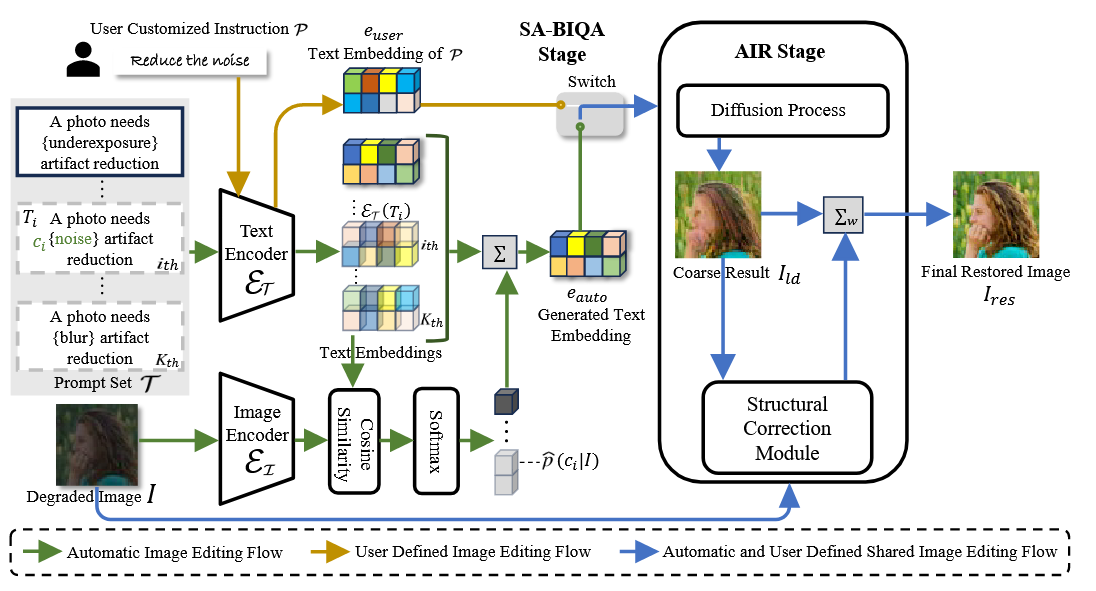}
   \includegraphics[width=.9\linewidth]{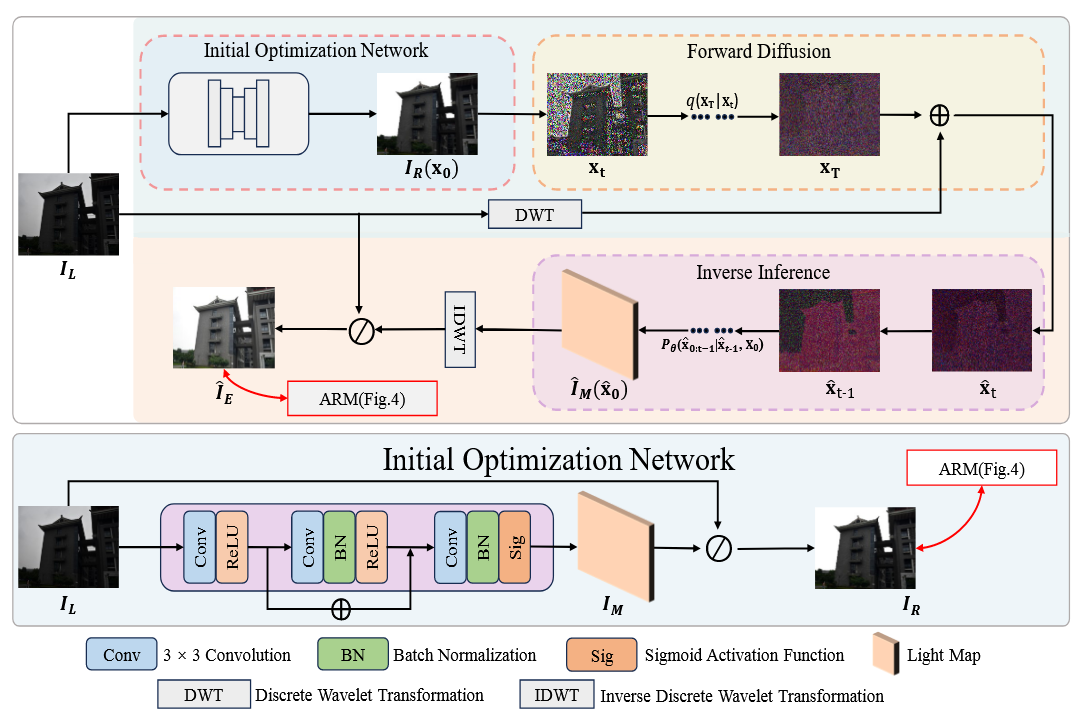}
   \caption{\textbf{[Top]}~The AutoDIR framework~\cite{jiang2024autodir} comprises SA-BIQA and AIR stages, where SA-BIQA automatically identifies degradations and generates text embeddings $e_{auto}$ via text encoder $\mathcal{E}_T$, while AIR employs a Structural Correction Module within the diffusion process to produce $I_{res}$. \textbf{[Bottom]}~The Zero-LED framework~\cite{zeroled2024} combines an Initial Optimization Network with bidirectional diffusion processes, where low-light input $I_L$ undergoes initial processing to generate $I_R(x_0)$, followed by forward diffusion $q(x_t|x_0)$ and reverse inference $p_\theta(x_{t-1}|x_t, x_0)$ to produce enhanced output $\hat{I}_E$.}
   \label{fig:autonomous_diffusion}
\end{figure}

    Zero-LED~\cite{he2024zeroLED}, as seen in the bottom portion of Figure~\ref{fig:autonomous_diffusion}, (previously discussed in Spectral Latent Diffusion) introduces a zero-reference learning approach, where the diffusion model learns to estimate lighting degradation without paired supervision. Instead of mapping low-light images to well-lit targets, Zero-LED aligns low-light images to the normal-light domain using frequency-domain constraints, ensuring adaptive and robust enhancement. ReDDiT~\cite{lan2024towards}, while primarily focused on acceleration, incorporates reflectance-aware self-supervision, allowing the model to learn from unpaired datasets by enforcing consistency in reflectance predictions across different lighting conditions. EWRD~\cite{ewrd} (Entropy-Weighted Reverse Diffusion) leverages an unsupervised entropy-based loss to guide the enhancement process. Instead of optimizing for pixel-wise differences, it learns image entropy distributions to enhance contrast and structure without paired data.

    % Takeaways 
    Zero-reference diffusion models eliminate the dependency on paired training data, making them more scalable to real-world conditions. However, their performance heavily relies on domain alignment strategies, and errors in estimating degradation distributions can lead to suboptimal enhancement quality.

\subsubsection{Unsupervised Domain Adaptation for Low-Light Enhancement}

    Since different imaging conditions result in dataset bias, models trained on specific lighting conditions often fail when deployed in real-world scenarios. Domain adaptation techniques help diffusion models transfer knowledge across different lighting conditions without requiring paired data.
    
    DiffLI$^2$D~\cite{yang2024unleashing} (discussed in Spectral Latent Diffusion) operates within a pre-trained diffusion model’s latent space, extracting semantic representations of low-light images and refining them without full retraining. By utilizing existing diffusion priors, it can generalize across diverse lighting conditions without the need for labeled low-light datasets. AutoDir~\cite{jiang2024autodir}, shown in the top of Figure~\ref{fig:autonomous_diffusion}, employs self-supervised domain alignment, dynamically selecting the most relevant modality (\textit{e.g.}, RGB, IR, or depth) for adaptation. This enables diffusion models to adjust their enhancement approach based on scene conditions, ensuring consistent performance across varied lighting environments.

    % Takeaways
    Unsupervised domain adaptation enhances generalization to unseen lighting conditions, making diffusion-based enhancement more reliable in real-world deployments. However, these models still face challenges in handling extreme lighting shifts, as domain alignment mechanisms may fail when degradation patterns are highly irregular.

    \begin{tcolorbox}[colback=gray!10, colframe=gray!40, sharp corners, boxrule=0.3pt, left=2pt, right=2pt, top=2pt, bottom=2pt]
    \textbf{Strengths:}
    \begin{itemize}[label=\cmark, itemsep=0.25pt, left=4pt]
        \item Eliminates the need for \textbf{paired training datasets}, improving scalability.
        \item Adaptable to \textbf{unseen lighting conditions}, increasing robustness.
        \item Self-supervised domain alignment improves \textbf{cross-environment generalization}.
    \end{itemize}

    \textbf{Limitations:}
    \begin{itemize}[label=\xmark, itemsep=0.25pt, left=4pt]
        \item Lacks explicit supervision, which can result in \textbf{suboptimal enhancement quality}.
        \item Requires \textbf{robust domain alignment}, which may fail in highly variable lighting conditions.
        \item Zero-reference models struggle with \textbf{severe distortions} where degradation estimation is inaccurate.
    \end{itemize}
    \end{tcolorbox}

    Table~\ref{tab:llie_taxonomy} summarizes the key aspects of the proposed taxonomy for diffusion-based LLIE methods:
    
    \begin{table*}[ht!]
\centering
\scriptsize
\renewcommand\arraystretch{1.0}
\setlength{\tabcolsep}{3pt}
\caption{\label{tab:llie_taxonomy}Summary of LLIE Diffusion Model Taxonomy}
\begingroup\sloppy
\resizebox{\textwidth}{!}{
    \begin{tabularx}{\textwidth}{L|L|L|L|L}
      \toprule[1.5pt]
      \textbf{Category}
        & \textbf{Core Idea/Principle}
        & \textbf{Representative SOTA Models (Examples)}
        & \textbf{\faCheckCircle~Key Strengths}
        & \textbf{\faExclamationTriangle~Key Limitations/Challenges} \\
      \midrule
      \rowcolor{LightCyan}\textbf{Intrinsic Decomposition} 
        & Decompose image into physical components (\textit{e.g.}, reflectance, illumination) for controlled enhancement.
        & \texttt{Diff-Retinex}, \texttt{LightenDiffusion}, \texttt{ExposureDiffusion} 
        & Improved interpretability, stable training, lower risk of hallucination.
        & Requires accurate prior estimation, higher computational cost, struggles when physical assumptions break down (\textit{e.g.}, non-uniform lighting). \\

      \textbf{Spectral \& Latent}
        & Perform diffusion in a transformed domain (frequency or learned latent space) for efficiency or targeted processing.
        & \texttt{SFDiff}, \texttt{DiffLL}, \texttt{L2DM}, \texttt{WaveDM}
        & Frequency: structural consistency, computational efficiency, scalability. Both: improved robustness to certain distortions.
        & Frequency: potential inverse transform artifacts. Latent: dependent on encoder quality, potential detail loss. \\

      \rowcolor{LightCyan}\textbf{Accelerated}
        & Reduce inference time of diffusion models via sampling reduction, distillation, or latent space operations.
        & \texttt{ReDDiT}, \texttt{PyDiff}, \texttt{L2DM}
        & Enables real-time/near real-time enhancement, makes diffusion models practical for resource-constrained devices.
        & Aggressive step reduction can cause blur/detail loss, distilled models may have reduced generalization, latent methods depend on encoder. \\

      \textbf{Guided}
        & Incorporate external signals (masks, text, parameters) to control the enhancement process spatially or semantically.
        & \texttt{CLE-Diffusion}, \texttt{InstructIR}, \texttt{ExposureDiffusion}
        & Spatially adaptive enhancement, user customizability, more natural brightness adjustments via exposure control.
        & Requires accurate guidance (masks, prompts), instruction models struggle with ambiguity, exposure models need robust lighting estimation. \\

      \rowcolor{LightCyan}\textbf{Multimodal}
        & Enhance for specific downstream tasks or fuse information from multiple sensor modalities (\textit{e.g.}, IR, event cameras).
        & \texttt{Diffusion-in-the-Dark}, \texttt{EVSNet}, \texttt{AutoDir}, \texttt{AGLLDiff}
        & Task-specific: improved downstream performance. Multi-modal: robust to extreme low-light where RGB fails, improved video consistency.
        & Higher computational complexity, limited multi-modal datasets, task-specific models lack broad generalization. \\

      \textbf{Autonomous}
        & Learn enhancement without paired data, using zero-shot, self-supervised, or unsupervised domain adaptation.
        & \texttt{Zero-LED}, \texttt{ReDDiT} (self-supervised part), \texttt{DiffLI2D}
        & Eliminates need for paired data (improves scalability), adaptable to unseen conditions, better cross-environment generalization.
        & Potentially suboptimal quality due to lack of direct supervision, relies on robust domain alignment which can fail. \\
      \bottomrule[1.5pt]
    \end{tabularx}
  }
  \endgroup
\end{table*}

%%%%%%%%%%%%%%%%%%%%%%%%%%%%%%%%%%%%%%%%%%%%%%%%%%%%%%%%%%%%%%%%
\section{Benchmark Datasets and Evaluation Metrics}\label{sec:datasets-metrics}
%%%%%%%%%%%%%%%%%%%%%%%%%%%%%%%%%%%%%%%%%%%%%%%%%%%%%%%%%%%%%%%%

    Evaluating the performance of low-light image enhancement (LLIE) methods, including those based on diffusion models, requires standardized datasets and appropriate evaluation metrics. This section details the commonly used benchmarks and metrics, discussing their characteristics, relevance, and limitations, particularly in the context of diffusion-based approaches. % TODO: ref for synthetic

    \subsection{Benchmark Datasets for LLIE}
    
    LLIE research relies on several key datasets for training and evaluation. Table~\ref{tab:datasets} summarizes the commonly used benchmarks, including their type (paired or unpaired), scale, and notable characteristics. Below we detail each dataset and its relevance:

    \begin{table}[ht!]
    \tiny 
    \centering
    \renewcommand\arraystretch{1.1}
    \caption{\label{tab:datasets}Common Benchmark Datasets for Low-Light Image Enhancement. Pairing: P = paired low/high images, U = unpaired. Scale: number of images (for paired datasets, number of pairs). Content: brief notes on content diversity or special features.}
    \setlength{\tabcolsep}{2pt}
    \resizebox{\textwidth}{!}{
    \begin{tabular}{p{1.5cm}lllll}
    \toprule[1.5pt]
    
    \textbf{Dataset} & \textbf{Type} & \textbf{Size} & \textbf{Scenes} & \textbf{Capture/Synthesis} & \textbf{Primary Usage} \\
    \hline
    \rowcolor{LightCyan}\textbf{LOL}~\cite{Wei2018RetinexNet} & Paired & 500 pairs & Mostly Indoor & Real capture (varying exposure/ISO) & Training, Testing \\
    \textbf{LOLv2}~\cite{yang2021} & Paired & Real: 789; Synth: 1000 & Indoor/Outdoor & Real capture \& Synthesis & Training, Testing \\
    \rowcolor{LightCyan}\textbf{LSRW} \cite{Hai2023R2RNet_LSRW} & Paired & 5,650 pairs & Diverse Indoor/Outdoor & Real capture (DSLR, smartphone), slight misalignment & Training, Testing (Generalization) \\
    \textbf{SID} \cite{Chen2018SID} & Paired (RAW) & 5,094 pairs (Sony + Fuji) & Indoor/Outdoor (Static, <0.1 lux) & Real capture (short vs long exposure, extreme low light) & Training, Testing (RAW-to-RGB) \\
    \rowcolor{LightCyan}\textbf{ExDark} \cite{Loh2019ExDark} & Unpaired & 7,363 images & Diverse (10 lighting conditions) & Real capture, 12 object classes annotated & Unsupervised LLIE, Task-based eval (Detection) \\
    \textbf{SICE} \cite{Cai2018SICE} & Multi-Exp Seq & 589 sequences ($\sim$4,800 images) & Diverse & Synthesized via MEF/HDR from sequences, some real & Training (Supervised/Unsupervised), Testing \\
    \rowcolor{LightCyan}\textbf{VE-LOL} \cite{Liu2021VELOL} & Paired (L) \& Unpaired (H) & L: 2.5k; H: $\sim$11k & Diverse (landscapes, urban, faces) & Real \& Synthetic, H has face annotations & Training, Testing (Perceptual \& Task-based) \\
    \textbf{NTIRE 2024} \cite{NTIRE2024challenge} & Paired (RAW) & 230 train, 70 val/test & Diverse, High-Res (4K+), Night & Real capture, non-uniform illumination, backlighting & Challenge Benchmark, SOTA evaluation \\
    \rowcolor{LightCyan}\textbf{MIT-Adobe FiveK} \cite{Bychkovsky2011FiveK} & Paired (RAW) & 5,000 images & Diverse & Real capture (DSLRs), expert retouched & Training (Global tonal adj.), Evaluation \\
    \textbf{ELD} \cite{Wei2020ELD} & Paired (RAW) & 10 scenes, 4 sensors & Indoor (Extreme low light) & Real capture (RAW, calibrated noise models) & Training, Testing (RAW Denoising) \\
    \rowcolor{LightCyan}\textbf{NEF} \cite{wang_naturalness_2013} & Unpaired & 156 images & Outdoor & Real capture (Canon) + web-sourced & Unsupervised LLIE \\
    \textbf{LLIE} \cite{shen_msr-netlow-light_2017} & Paired & 10,000 pairs & Diverse & Manually-corrected, web-sourced w/ gamma corrections & Training, Testing \\
    \rowcolor{LightCyan}\textbf{DARKZURICH} \cite{sakaridis_guided_2019} & Paired (day/night) & Day: 3,041; Night: 2,416 & Outdoor & Real capture (GoPro Hero 5, 1080p) & Road-scene parsing, Semantic annotation \\
    \rowcolor{LightCyan}\textbf{RELLISUR} \cite{aakerberg_rellisur_2021} & Paired & 12,750 pairs (3 scales) & Diverse (50/50 indoor/outdoor) & Real capture (Canon EOS 6D, multiple scales) & Training, Testing; Low-light \& low-resolution \\
    \textbf{LRD} \cite{zhang_towards_2023} & Paired (RAW) & 1,800 images (100×6×3) & Diverse & Real capture (IMX586 sensor) & Training, Testing, Low-light RAW denoising \\
    \rowcolor{LightCyan}\textbf{UHD-LOL} \cite{wang_ultra-high-definition_2023} & Paired & 11,065 pairs (4K/8K) & Diverse & Web-sourced, Adobe Lightroom altered & Training, Testing, Ultra-HD LLIE \\
    \textbf{FEW-SHOTS} \cite{prabhakar_few-shot_2023} & Paired (RAW) & 194 images (65 pairs) & Diverse & Short/long exposure (Nikon D5600) & Training, Testing, Low-light RAW \\
    \rowcolor{LightCyan}\textbf{LOM} \cite{cui_aleth-nerf_2023} & Paired & 25-65 pairs/scene & 5 real-world scenes & DJI Osmo Action 3 (various exp/ISO) & Training, Testing, sRGB LLIE, Multi-view \\
    \textbf{LENVIZ} \cite{aithal_lenviz_2025} & Paired & 234,688 frames (24,082 scenes) & Diverse & Multi-exposure frames (3 MOS sensors) & Training, Testing, Pixel-level, HDR \\

    \bottomrule[1.5pt]
    \end{tabular}%
    }
\end{table}

    \begin{itemize}[leftmargin=*,itemsep=0pt,topsep=2pt]
        \item \textbf{LOL (LOw-Light)} – Introduced by Wei \emph{et al.} \cite{Wei2018RetinexNet} for RetinexNet, LOL is a paired dataset of 500 images (485 training pairs, 15 testing) of real scenes captured in low/high light conditions. It has limited scene diversity (mostly indoor) and relatively low resolution, which makes it less representative of complex real-world low-light scenarios. The small test set often leads to overfitting on this benchmark. This motivated the creation of larger datasets addressing LOL’s limitations (\textit{e.g.} monotonic content and uniform illumination). LOLv2 expanded this with real ($\sim$789 pairs) and synthetic ($\sim$1000 pairs) subsets~\cite{yang2021}.

        \item \textbf{LSRW (Large-Scale Real-World)} – Hai \emph{et al.} \cite{Hai2023R2RNet_LSRW} built LSRW, the first large-scale real-world paired dataset with 5,650 low/normal-light image pairs. It consists of diverse indoor and outdoor scenes captured with a Nikon D7500 DSLR (3,170 pairs) and a Huawei P40 smartphone (2,480 pairs). The images have slight misalignment due to hand-held capture, reflecting practical conditions. LSRW’s scale and real pairing improve model generalization to real low-light inputs.

        \item \textbf{SID (See-in-the-Dark)} – Chen \emph{et al.} \cite{Chen2018SID} collected the Sony/Fuji SID dataset of short-exposure RAW images with corresponding long-exposure reference shots. It contains 5,094 paired images (split 7:1:2 for train/val/test) from extremely dark indoor (0.03–0.3 lux) and outdoor (0.2–5 lux) scenes. SID is focused on raw sensor data and extreme low-light noise, making it popular for raw-denoising and enhancement tasks. It provides high-resolution (~4K) pairs but is limited to specific cameras and scenes.

        \item \textbf{ExDark (Exclusively Dark)} – Loh and Chan \cite{Loh2019ExDark} introduced ExDark, an unpaired dataset of 7,363 low-light images across 10 lighting conditions (\textit{e.g.} candle, moonlight, twilight). Images cover diverse indoor and outdoor scenes with 12 object classes (similar to PASCAL VOC) annotated for object detection. ExDark is split into 4,800 train and 2,563 test images. While not providing normal-light references for enhancement, ExDark is valuable for evaluating enhancement impact on detection tasks and for unsupervised or domain adaptation methods. Its diversity and object annotations make it relevant for task-driven LLIE evaluations.

        \item \textbf{SICE (Single-Image Contrast Enhancement)} – Cai \emph{et al.} \cite{Cai2018SICE} constructed SICE from multi-exposure sequences to cover under- and over-exposed conditions. SICE contains 4,413 images from 589 scenes, including diverse indoor/outdoor content. Each scene has multiple exposures and a high-quality fused image as reference. A typical split uses 360 sequences (Part 1) for training and 229 (Part 2) for testing, totaling about 4,800 images. Uniquely, SICE contains both low-light and normal/over-light images, enabling supervised training or unsupervised/zero-shot settings by treating over-exposed images as additional references. However, some images are synthetically fused and not strictly real photographs. Recent variants (SICE\_Grad, SICE\_Mix) further extend this dataset to include more extreme exposures.

        \item \textbf{VE-LOL (Vision Enhancement in Low Light)} – Liu \emph{et al.} \cite{Liu2021VELOL} presented VE-LOL to overcome LOL’s shortcomings. VE-LOL is a large-scale dataset with 13,440 images covering diverse scenes (natural landscapes, urban streets, human faces, etc.). It comprises two subsets: VE-LOL-L (paired subset with 2,100 low/normal pairs for training and 400 for testing) and VE-LOL-H (unpaired high-level subset with 6,940 low-light images for training and 4,000 for testing). VE-LOL-H also provides face detection bounding box annotations, enabling joint low-light enhancement and detection benchmarking. The inclusion of real-captured and synthetic low-light images, higher resolution content, and annotated faces makes VE-LOL a comprehensive benchmark for both human vision (perceptual quality) and machine vision (detection performance) in low-light. It addressed the limited scenes and lack of high-level task consideration in older datasets.

        \item \textbf{NTIRE 2024 LLIE} – The NTIRE 2024 challenge \cite{NTIRE2024challenge} introduced a new LLIE dataset to push current limits. It features a wide range of scenes under various low-light conditions, from dim indoor environments to extremely dark outdoor night scenes. Unlike prior datasets, it includes high-resolution images (up to 4K) and more night scene content to reflect modern smartphone photography challenges. Specifically, the NTIRE2024 dataset has 230 training scenes and 70 validation/testing scenes (ground truths hidden for the challenge) covering non-uniform illumination, backlit scenes, and other difficult lighting cases. This dataset was created because earlier datasets like LOL and FiveK have “limited resolution, monotonous content, and uniform illumination levels”, which hamper the performance of cutting-edge diffusion and enhancement models on real-world photos. By greatly increasing content diversity and resolution, the NTIRE 2024 dataset serves as a new benchmark for evaluating LLIE, especially for diffusion-based and high-capacity models.

        \item \textbf{MIT-Adobe FiveK} – Bychkovsky \emph{et al.} \cite{Bychkovsky2011FiveK} collected 5,000 raw photographs taken by DSLR cameras, each retouched by five expert photographers for exposure and color enhancements. While not exclusively low-light, FiveK spans a broad range of lighting conditions and is often used for learning global tonal adjustment and enhancement. In LLIE works, FiveK (usually using one expert’s retouched version as ground truth) provides diverse, high-quality reference images for training or evaluation. It is unpaired in the sense that input images vary in lighting (some underexposed, some normal) and the outputs are the expert-adjusted versions. FiveK’s diversity and professional adjustments make it a strong benchmark for color and exposure enhancement models beyond extreme low-light cases.

        \item \textbf{ELD (Extreme Low-light Denoising)} – Wei \emph{et al.} \cite{Wei2020ELD} introduced the ELD dataset to study noise in very dark conditions. ELD includes images of 10 indoor scenes captured with 4 different camera sensors (Sony, Nikon, Canon, etc.), with extremely low exposures and corresponding long-exposure reference images. It provides RAW sensor data with accurate calibration (including dark frames) and a physics-based noise model. While primarily aimed at denoising, ELD is relevant to LLIE as it benchmarks how enhancement models handle real noise from different devices. The dataset’s multi-camera, real noise nature addresses the sensor-specific bias of SID (which had only two cameras) and helps evaluate model robustness across hardware. Generative diffusion models for LLIE, which often synthesize detail and noise, can be evaluated on ELD to ensure they produce realistic noise-free results in extremely dark conditions.
    \end{itemize}
    
    In addition to the above, several legacy datasets are occasionally used for evaluation, though they are smaller in scale and often lack ground truth. These include DICM, LIME, MEF, NPE, and VV collections (each containing just a few dozen low-light images). These served as early benchmarks for traditional enhancement methods but lack ground-truth references and diversity. 
    Similarly, task-specific datasets like Dark Face \cite{yang2021face} and ACDC \cite{sakaridis2021acdc} focus on nighttime detection (with thousands of images for face detection and driving scenes respectively) and are used when evaluating task-driven enhancement.
    
    Earlier datasets like LOL and small-scale collections proved useful to kickstart research, but their limitations (few samples, simple scenes, mostly static setups, limited resolution) became apparent as models grew more powerful. For instance, models trained only on LOL often do not generalize well to real photographs due to the narrow data distribution. This spurred the development of newer benchmarks: VE-LOL addressed the lack of high-level context and scene variety by adding human-centric scenes and more samples; LSRW and NTIRE 2024 tackled the need for high-resolution, real-captured data in diverse settings to train modern diffusion models; and SICE~\cite{Cai2018SICE} included both under- and over-exposed images to broaden the enhancement scope. In summary, the evolution of LLIE datasets has moved toward larger-scale, more diverse, higher-resolution, and task-oriented collections, which better reflect the challenges of low-light conditions in the wild and support the training and evaluation of diffusion-based enhancers under more realistic conditions.

\subsection{Evaluation Metrics for LLIE}

    Evaluating low-light enhancement results requires a multifaceted approach. Researchers employ a mix of full-reference metrics (when ground truth enhanced images are available) and no-reference metrics (when only the input/output is given), as well as perceptual and task-specific measures. Table~\ref{tab:metrics} categorizes common metrics by their type, along with a brief description. We outline each category and discuss their trade-offs, especially in the context of generative diffusion models.

    \begin{table}[ht!]
\scriptsize
\centering
\renewcommand\arraystretch{1.2}
\caption{\label{tab:metrics}Common Evaluation Metrics for Low-Light Image Enhancement. FR = full-reference, NR = no-reference, Dist. = distribution-based.\vspace{0.25cm}}
\setlength{\tabcolsep}{3pt}
    \begin{tabular}{p{5.5cm}ccp{7cm}}
    \toprule[1.25pt]
    \textbf{Metric Name} & \textbf{Abbrev.} & \textbf{Type} & \textbf{Description} \\
    \hline
    \rowcolor{LightCyan}\textbf{Peak Signal-to-Noise Ratio} & PSNR & FR (pixel fidelity) & Measures pixel-wise reconstruction fidelity based on Mean Squared Error (MSE). Higher is better. \\
    \textbf{Structural Similarity Index Measure} \cite{Wang2004SSIM} & SSIM & FR (structural) & Measures similarity based on luminance, contrast, and structure. Value range [0, 1], higher is better. \\
    \rowcolor{LightCyan}\textbf{Learned Perceptual Image Patch Similarity} \cite{Zhang2018LPIPS} & LPIPS & FR (perceptual) & Measures perceptual similarity using deep features (\textit{e.g.}, VGG). Lower is better (more similar). \\
    \textbf{Natural Image Quality Evaluator} \cite{mittal2012niqe} & NIQE & NR (quality) & Blind metric assessing ``naturalness'' based on statistical features compared to a natural scene model. Lower is better. \\
    \rowcolor{LightCyan}\textbf{Perceptual Index} \cite{Blau2018PIRM} & PI & NR (quality) & Combines NIQE and other factors (\textit{e.g.}, Ma score) for perceptual quality assessment. Lower is better. \\
    \textbf{Fréchet Inception Distance} \cite{Heusel2017FID} & FID & Dist. (realism) & Measures distance between feature distributions of generated and real images (using Inception network). Lower is better. \\
    \rowcolor{LightCyan}\textbf{Kernel Inception Distance} \cite{Binkowski2018KID} & KID & Dist. (realism) & Similar to FID but uses a polynomial kernel (Maximum Mean Discrepancy), potentially more robust for smaller sample sizes. Lower is better. \\
    \textbf{Deep Image Structure and Texture Similarity} \cite{Wang2004SSIM} & DISTS & FR (perceptual) & Perceptual metric considering structure and texture similarity using deep features. Lower is better. \\
    \rowcolor{LightCyan}\textbf{Task-Based Evaluation} (\textit{e.g.}, mAP/Acc) & - & Application / Task-based & Evaluating the impact of LLIE on downstream tasks like object/face detection (mAP) or segmentation/classification (Accuracy). Higher is better. \\
    \bottomrule[1.25pt]
    \end{tabular}
\end{table}

    \subsubsection{Full-Reference Metrics (FR)}
    These metrics compare the enhanced image to a ground truth reference. \textbf{PSNR} (Peak Signal-to-Noise Ratio) is a pixel-wise fidelity measure in decibels; higher PSNR indicates the result is very similar to the reference in terms of raw error. \textbf{SSIM} (Structural Similarity Index) \cite{Wang2004SSIM} measures perceptual structural similarity (luminance, contrast, and structure) between images, often correlating better with human perception than PSNR. Both PSNR and SSIM are default metrics in LLIE papers to report reconstruction quality. However, they primarily reward fidelity and may penalize legitimate enhancements that deviate from the reference exposure or color. \textbf{LPIPS} (Learned Perceptual Image Patch Similarity) \cite{Zhang2018LPIPS} is also full-reference but uses deep network features to judge similarity. LPIPS (lower is better) reflects perceptual closeness by accounting for human-sensitive differences (textures, edges) rather than pixel-wise error. 
    \textbf{DISTS} \cite{lan2024towards} is another perceptual FR metric that considers structure and texture similarity.
    These perceptual metrics (LPIPS, DISTS) are helpful when methods produce outputs that are visually plausible but not pixel-identical to the ground truth (as often the case with GAN or diffusion models).

    \subsubsection{No-Reference Metrics (NR)}
    In many cases (especially with real night photos or unpaired datasets), no exact ground truth is available, so blind Image Quality Assessment (IQA) metrics are used. 
    \textbf{NIQE} (Natural Image Quality Evaluator) \cite{mittal2012niqe} is an opinion-unaware metric that measures deviations from statistical naturalness in an image.
    Lower NIQE indicates the image’s statistics are closer to those of high-quality natural images. NIQE is popular in LLIE to assess quality without reference, but it may not always correlate well with human opinion for enhanced images. The \textbf{Perceptual Index (PI)} \cite{Blau2018PIRM} combines NR measures into one score, often defined as $\text{PI} = \frac{1}{2}((10 - \text{Ma}) + \text{NIQE})$, where Ma is a learned no-reference quality score. Lower PI means better perceptual quality. LLIE works have adopted PI to quantify the trade-off between making images look good versus staying faithful. It is worth noting that NR metrics can sometimes be gamed or fail on very out-of-distribution images, so their scores must be interpreted with caution.

    \subsubsection{Distribution-Based Metrics}
    Generative diffusion models produce outputs intended to match the distribution of clean, well-exposed images. To evaluate this realism at the distribution level, metrics like \textbf{FID} (Fréchet Inception Distance) \cite{Heusel2017FID} and \textbf{KID} (Kernel Inception Distance) \cite{Binkowski2018KID} are employed. FID computes the distance between multivariate Gaussians fit to deep feature embeddings (\textit{e.g.}, InceptionV3 features) of the enhanced images and reference “good” images. Lower FID indicates the enhanced images collectively have feature statistics close to real images (lower means more realistic and diverse). KID is a related metric based on the Maximum Mean Discrepancy (MMD) in feature space; it is an unbiased alternative to FID, potentially more robust for smaller sample sizes. Both FID and KID are used to evaluate GANs and diffusion models, and they have started appearing in LLIE evaluations when the goal is to produce photo-realistic enhancements rather than exactly replicating ground truth. These metrics do not require one-to-one image correspondences but require a set of enhanced images and a set of target reference images.

    \subsubsection{Task-Based Metrics}
    Ultimately, the value of enhancement can be measured by its impact on downstream tasks. Task-based evaluation uses performance metrics from applications like detection or recognition on the enhanced images. For instance, on the VE-LOL dataset, enhancement methods are evaluated by the accuracy of face detection on the brightened images. Metrics include detection Average Precision (AP) or F1-score for object/face detection (\textit{e.g.}, using ExDark~\cite{Loh2019ExDark} or Dark Face datasets), classification accuracy in low-light conditions, or segmentation mIoU on datasets like ACDC. Task-based metrics treat the enhancement algorithm as part of a pipeline and assess whether the overall system’s performance improves (\textit{e.g.}, higher mAP for detecting pedestrians at night after using the enhancer). These metrics are especially pertinent for diffusion models, which can be tuned to not only produce visually pleasing results but also ones that retain features important for machine vision.

    \subsubsection{Trade-offs and Considerations}
    Traditional metrics like PSNR/SSIM focus on low-level fidelity and often fail to reflect perceptual quality adequately. Maximizing PSNR can lead to blurred or over-smoothed outputs. Optimizing purely for perceptual quality (minimizing LPIPS/DISTS or NIQE, or improving FID/KID) can produce images that diverge from the original structure or colors if the model is over-regularized. Diffusion models, being generative, tend to prioritize producing realistic lighting, textures, and noise patterns – often at the expense of exact pixel reproduction. As a result, diffusion-based LLIE methods might score lower on PSNR than deterministic regression models, yet yield more visually pleasing or authentic results (reflected in better PI, lower FID, or superior human opinion scores). Researchers therefore use a combination of metrics to get a holistic evaluation. A high PSNR/SSIM ensures the method is not hallucinating incorrect content, while good perceptual scores (LPIPS, DISTS, PI) or distribution metrics (FID, KID) indicate the enhancement looks natural and detailed. The best diffusion-based LLIE approaches balance pixel-level fidelity with perceptual realism and downstream utility.

%%%%%%%%%%%%%%%%%%%%%%%%%%%%%%%
\subsection{Cross-Benchmark Performance Landscape}
\label{subsec:metrics}
%%%%%%%%%%%%%%%%%%%%%%%%%%%%%%%

    This section provides a quantitative and qualitative comparison of state-of-the-art (SOTA) diffusion model-based LLIE methods against other prominent generative approaches such as Transformers, GANs, Deep CNNs. The objective is to offer a clear overview of the current performance landscape, highlighting trends, strengths, and weaknesses of different model families on established benchmarks. The comparison considers a suite of metrics including PSNR, SSIM, LPIPS, alongside model complexity indicators such as the number of parameters and FLOPs. 
    Datasets like LOL (v1, v2-real, v2-synthetic)~\cite{Wei2018RetinexNet, yang2021}, SID~\cite{Chen2018SID}, VE-LOL~\cite{Liu2021VELOL}, and results from NTIRE challenges~\cite{NTIRE2024challenge} are central to this analysis due to their common usage and the specific LLIE challenges they present. % TODO: ref for synthetic

    \begin{table}[htbp]
\centering
\caption{\label{tab:llie_sota_improved}Performance Comparison of Low-Light Image Enhancement Models. Method categories are abbreviated: Diffusion Models (DM), Transformers (TF), and CNN/Hybrid Methods (DCNN). For LOLv2 results, values formatted as $xx/xx$ correspond to scores on Real/Synthetic datasets, respectively.}

\resizebox{\columnwidth}{!}{

    \begin{tabular}{@{}llllcccccccr@{}}
    \toprule[1.5pt]

    \multirow{2}{*}{\textbf{\begin{tabular}[c]{@{}c@{}}Method\\Category\end{tabular}}} & 
    \multirow{2}{*}{\textbf{Model Name}} & 
    \multirow{2}{*}{\textbf{Year}} & 
    \multirow{2}{*}{\textbf{Venue}} & 
    \multicolumn{3}{c}{\textbf{LOLv1}} & 
    \multicolumn{3}{c}{\textbf{LOLv2}} & 
    \multirow{2}{*}{\textbf{Params (M)}} \\

    \cmidrule(lr){5-7} \cmidrule(lr){8-10}
    & & & & 
    \textbf{PSNR} & \textbf{SSIM} & \textbf{LPIPS} & \textbf{PSNR} & \textbf{SSIM} & \textbf{LPIPS} & \\ \midrule
    
    \rowcolor{LightCyan}DM & 
    GSAD~\cite{hou2023global} & 
    2023 & NeurIPS & 
    27.83 & 0.877 & 0.091 & 
    28.81/28.67 & 0.895/0.944 & 0.095/0.047 & 
    17.40 \\
    
    DM & 
    PyDiff~\cite{pydiff2023} &
    2023 & arXiv & 
    27.09 & 0.93 & 0.10 & 
    24.01 & 0.876 & 0.23 & 
    97.90 \\

    \rowcolor{LightCyan}DM & 
    DiffLLE~\cite{yang2023difflle} & 
    2023 & TOG & 
    26.34 & 0.845 & - & 
    - & - & - & 
    22.10 \\

    DM & 
    ReDDit~\cite{lan2024towards} & 
    2025 & CVPR &
    27.98 & 0.872 & 0.053 & 
    31.25/30.03 & 0.904/0.933 & 0.038/0.031 & 
    18.00 \\

    DM & 
    $\text{MFF-Diffusion}_{\text{LLformer}}$~\cite{shi_multi-feature_2025} & 
    2025 & MDPI & 
    23.75 & 0.872 & 0.065 & 
    23.56 & 0.862 & 0.161 & 
    97.39 \\

    \rowcolor{LightCyan}DM & 
    AdaptDiff~\cite{shao_adaptdiff_2025} & 
    2025 & CVGIP: Image Understanding & 
    27.28 & 0.907 & - & 
    27.59/28.17 & 0.895/0.943 & - & 
    18.47 \\

    DM & 
    SVD-GDiff~\cite{kim_svd-guided_2025} & 
    2025 & IEEE Signal Process. Lett & 
    18.71 & 0.65 & 0.276 & 
    18.86 & 0.650 & 0.272 & 
    - \\

    \rowcolor{LightCyan}DM & 
    Bidirectional Diff~\cite{he_degradation-consistent_2025} &
    2025 & ACM MM & 
    26.95 & 0.869 & 0.164 & 
    23.37/28.88 & 0.875/0.953 & 0.181/0.049 &
    33.50 \\

    DM & 
    DMCDiff~\cite{wu_dmcdiff_nodate} & 
    2025 & SSRN & 
    24.40 & 0.87 & 0.14 & 
    -/23.71 & -/0.91 & -/0.09 & 
    - \\

    \rowcolor{LightCyan}DM & 
    Diff-Retinex~\cite{diff-retinex} & 
    2023 & ICCV & 
    22.69 & 0.853 & 0.191 & 
    21.19/24.33 & 0.833/0.921 & 0.271/0.061 & 
    56.90 \\
    
    DM & 
    CLE-Diffusion~\cite{yin2023cle} &
    2023 & ACM MM & 
    25.51 & 0.89 & 0.16 & 
    - & - & - & 
    37.01 \\
    
    \rowcolor{LightCyan}DM &
    LLDiffusion~\cite{wang2023lldiffusion} &
    2025 & Pattern Recognition & 
    24.65 & 0.843 & 0.075 & 
    23.16/25.99 & 0.842/0.948 & - & 
    - \\
    
    DM & 
    GSAD~\cite{hou2023global} &
    2023 & NeurIPS & 
    23.22 & 0.855 & 0.193 & 
    20.18/24.10 & 0.847/0.927 & 0.205/0.073 &
    17.40 \\
    
    \rowcolor{LightCyan}DM &
    Entropy-SDE~\cite{entropy-sde} & 
    2024 & CVPR & 
    24.05 & 0.848 & 0.081 & 
    21.31 & 0.832 & 0.120 & 
    - \\
    
    DM & 
    LightenDiffusion~\cite{lightendiffusion} & 
    2024 & ECCV & 
    20.19 & 0.814 & 0.316 & 
    21.10/21.54 & 0.847/0.866 & 0.305/0.188 &
    27.80 \\
    
    \rowcolor{LightCyan}DM & 
    Diffusion in the Dark~\cite{nguyen2024diffusion} &
    2024 & CVPR & 
    23.97 & 0.84 & 0.12 &
    - & - & - & 
    55.70 \\
    
    DM & 
    L2DM~\cite{L2DM} &
    2023 & PRCV & 
    24.54 & 0.83 & - & 
    24.80/23.96 & 0.817/0.786 & - &
    0.07 \\
    
    \rowcolor{LightCyan}DM & 
    SFDiff~\cite{wan2024sfdiff} &
    2024 & IET Image Processing & 
    23.70 & 0.833 & 0.152 & 
    21.71/25.89 & 0.829/0.924 & 0.157/0.085 & 
    23.08 \\
    
    DM & 
    ZeroLED~\cite{zeroled2024} & 
    2024 & arXiv & 
    19.82 & 0.841 & 0.242 &
    19.95 & 0.781 & 0.272 & 
    - \\

    \rowcolor{LightCyan}DM &
    InstructIR~\cite{conde2024instructir} &
    2024 & ECVA &
    22.81 & 0.836 & - &
    24.95 & 0.872 & - & 
    15.80 \\
    
    DM & 
    DiffLI2D~\cite{yang2024unleashing} &
    2024 & ECVA &
    23.30 & 0.849 & 0.136 &
    22.35 & 0.874 & 0.186 &
    8.63 \\
    
    \rowcolor{LightCyan}DM & 
    MDMS~\cite{mdms} & 2024 & AAAI &
    27.12 & 0.882 & 0.078 &
    33.30/17.40 & 0.933/0.797 & 0.043/0.227 &
    - \\

    DM & 
    AGLLDiff~\cite{lin2024aglldiff} &
    2025 & AAAI &
    21.81 & 0.84 & 0.15 &
    -/21.11 & -/0.87 & -/0.13 &
    - \\
    
    \rowcolor{LightCyan}DM & 
    RetiDiff~\cite{he2023retidiff} & 
    2025 & ICLR & 
    25.13 & 0.866 & 0.199 &
    22.99/27.53 & 0.859/0.951 & 0.198/0.053 &
    26.10 \\
    
    DM & 
    FourierDiff~\cite{Lv_2024_CVPR} & 
    2024 & CVPR & 
    17.56 & 0.607 & 0.359 & 
    18.67/13.70 & 0.602/0.631 & 0.362/0.252 & 
    547.50 \\
    
    \rowcolor{LightCyan}DM & 
    DiffLight~\cite{Feng_2024_CVPR} &
    2024 & CVPR & 
    25.85 & 0.876 & 0.082 &
    - & - & - & 
    51.70 \\
    
    DM & 
    MDDE~\cite{MDDE} & 
    2024 & ICDSP &
    20.93 & 0.828 & - &
    - & - & - & 
    - \\
    
    \rowcolor{LightCyan}DM &
    PSC Diffusion~\cite{wan2024psc} & 
    2024 & ACM Multimedia Systems &
    19.46 & 0.884 & 0.102 &
    21.13 & 0.876 & 0.121 & 
    67.70 \\
    
    DM & 
    JoRes-Diff~\cite{wu2024jores} & 
    2024 & ACM MM & 
    26.49 & 0.876 & 0.092 & 
    24.84/25.71 & 0.897/0.928 & 0.109/0.058 & 
    - \\
    
    \rowcolor{LightCyan}DM &
    SG-DDPM~\cite{wang2024sg} & 
    2024 & ICCEDE & 
    25.01 & 0.865 & - & 
    21.91 & 0.877 & - & 
    - \\
    
    DM & 
    Palette~\cite{saharia2022palette} & 
    2022 & SIGGRAPH &
    11.78 & 0.561 & 0.498 &
    14.71 & 0.692 & 0.333 &
    - \\
    
    \rowcolor{LightCyan}DM & 
    TDS~\cite{yang2020fidelity} & 
    2023 & CSECS &
    24.85 & 0.89 & 0.11 &
    - & - & - &
    - \\ \midrule\midrule

    \rowcolor{LightCyan}GANs & 
    EnlightenGAN~\cite{jiang2021enlightengan} &
    2021 & TIP & 
    17.48 & 0.652 & 0.322 & 
    18.64/16.57 & 0.677/0.734 & 0.309 & 
    8.64 \\ 

    GANs & 
    LE-GAN~\cite{fu2022gan} &
    2022 & Elsevier KBS & 
    - & - & - & 
    22.45 & 0.886 & - & 
    9.91 \\ 
    
    % Transformer models
    TF & 
    Uformer~\cite{wang2021uformer} &
    2022 & CVPR &
    19.00 & 0.741 & 0.354 &
    18.44 & 0.759 & 0.347 & 
    50.90 \\

    \rowcolor{LightCyan}TF & 
    Restormer~\cite{zamir2021restormer} &
    2022 & CVPR &
    20.61 & 0.797 & 0.288 &
    24.91 & 0.851 & 0.264 &
    26.10 \\
    
    TF &
    LLformer~\cite{wang2023ultra} &
    2023 & AAAI &
    22.89 & 0.816 & 0.202 &
    23.13 & 0.855 & 0.153 &
    1.61 \\
    
    \rowcolor{LightCyan}TF &
    Retinexformer~\cite{cai2023retinexformer} &
    2023 & ICCV &
    25.16 & 0.845 & 0.190  &
    22.80/25.67 & 0.840/0.930 & 0.210/0.115  &
    1.61 \\
    
    TF &
    IAT~\cite{Cui_2022_BMVC} &
    2022 & BMVC &
    23.38 & 0.809 & 0.240  &
    23.50 & 0.824 & 0.210  & 
    0.09 \\
    
    \rowcolor{LightCyan}TF &
    LYT-Net~\cite{brateanu2025lyt} &
    2024 & IEEE Signal Process. Lett &
    27.23 & 0.853 & 0.083 &
    27.80 & 0.873 & 0.071 &
    0.045 \\
    
    TF & 
    U-EGformer~\cite{adhikarla2024unified} &
    2024 & ICPR &
    23.56 & 0.836 & 0.225 
    &
    22.05 & 0.841 & 0.230 
    &
    0.10 \\
    
    \rowcolor{LightCyan}Mamba & 
    RetinexMamba~\cite{bai2024retinexmamba} & 
    2024 & ICONIP & 
    24.02 & 0.827 & 0.240 
    & 
    22.45 & 0.844 & 0.220 
    & 
    - \\
    
    Mamba &
    FourierTMamba~\cite{peng2024low} &
    2024 & ACM MM & 
    24.33 & 0.845 & 0.235 
    & 
    22.41 & 0.860 & 0.225 
    & 
    - \\
    
    \rowcolor{LightCyan}Mamba &
    WaveletMamba~\cite{shang2025waveletmamba} &
    2025 & Springer Display &
    23.27 & 0.851 & 0.228 
    & 
    22.49 & 0.869 & 0.218 
    & 
    - \\
    
    Mamba & 
    ExpoMamba~\cite{adhikarla2024expomamba} &
    2024 & ICML (ESFoMo-II) &
    25.77 & 0.860 & 0.180 
    & 
    28.04 & 0.885 & 0.160 
    &
    41.00 \\
    
    % DCNN models (bottom part)
    
    DCNN &
    Retinex~\cite{wei2018deep} &
    2018 & BMVC &
    16.77 & 0.462 & 0.417 &
    17.72 & 0.652 & 0.436 &
    0.45 \\
    
    \rowcolor{LightCyan}DCNN &
    KinD~\cite{zhang2019kindling} &
    2019 & ACM MM &
    17.65 & 0.771 & 0.175 &
    19.20 & 0.800 & 0.310 
    & 
    0.59 \\
    
    DCNN &
    DSLR~\cite{lim2020dslr} &
    2020 & TMM &
    14.82 & 0.572 & 0.375 &
    17.00 & 0.596 & 0.408 &
    0.60 \\
    
    \rowcolor{LightCyan}DCNN &
    DRBN~\cite{yang2020fidelity} &
    2020 & CVPR &
    16.68 & 0.73 & 0.345 &
    18.47 & 0.768 & 0.352 &
    0.53 \\
    
    DCNN &
    Zero-DCE~\cite{guo2020zero} &
    2020 & CVPR &
    14.86 & 0.562 & 0.372 &
    18.06 & 0.58 & 0.352 &
    0.079 \\
    
    \rowcolor{LightCyan}DCNN &
    MIRNet~\cite{zamir2020learning} &
    2020 & ECCV &
    24.14 & 0.830 & 0.250 &
    20.02 & 0.82 & 0.233 &
    31.80 \\
    
    DCNN &
    Zero-DCE++~\cite{li2021learning} &
    2021 & TPAMI &
    14.75 & 0.517 & 0.328 &
    17.90 & 0.560 & 0.340 
    &
    0.01 \\
    
    \rowcolor{LightCyan}DCNN &
    ReLLIE~\cite{zhang2021rellie} & 
    2021 & ACM MM &
    11.44 & 0.482 & 0.375 &
    14.40 & 0.536 & 0.334 &
    0.80 \\
    
    DCNN &
    RUAS~\cite{liu2021benchmarking} &
    2021 & CVPR &
    16.41 & 0.503 & 0.364 &
    15.35 & 0.495 & 0.395 &
    0.00 \\
    
    \rowcolor{LightCyan}DCNN &
    SCI~\cite{ma2022toward} &
    2022 & CVPR &
    14.78 & 0.525 & 0.366 &
    17.30 & 0.54 & 0.345 &
    0.00 \\
    
    DCNN &
    URetinex-Net~\cite{wu2022uretinex} &
    2022 & CVPR &
    19.84 & 0.824 & 0.237 &
    21.09 & 0.858 & 0.208 &
    1.32 \\
    
    \rowcolor{LightCyan}DCNN &
    SNRNet~\cite{xu2022snr} &
    2022 & CVPR & 
    23.43 & 0.843 & 0.234 &
    21.48 & 0.849 & 0.237 &
    5.20 \\
    
    DCNN &
    CDEF~\cite{valin2016daala} &
    2022 & TMM &
    16.34 & 0.585 & 0.407 &
    19.76 & 0.63 & 0.349 &
    2.10 \\
    
    \rowcolor{LightCyan}DCNN &
    UHDFour~\cite{li2023embedding} &
    2023 & ICLR &
    23.09 & 0.821 & 0.259 &
    21.79 & 0.854 & 0.292 &
    22.50 \\
    
    DCNN &
    CIDNet~\cite{Verma_2025_CVPR} &
    2024 & CVPR &
    28.14 & 0.889 & 0.079 &
    23.90 & 0.865 & 0.121 & 
    1.88 \\
    
    \rowcolor{LightCyan}DCNN &
    CFWD~\cite{DBLP:journals/corr/abs-2401-03788} &
    2024 & CoRR &
    29.19 & 0.872 & 0.197 &
    29.85 & 0.891 & 0.193 & 
    34.00 
    \\
    
    DCNN &
    BEM~\cite{huang2025bayesianneuralnetworksonetomany} &
    2025 & ICLR &
    28.80 & 0.884 & 0.069 &
    24.20 & 0.860 & 0.120 &
    28.00 
    \\

    \bottomrule[1.5pt]
\end{tabular}}
\end{table}

    \textbf{Discussion of Quantitative and Qualitative Results}
    
    The quantitative results presented in Table~\ref{tab:llie_sota_improved}, synthesized from numerous recent papers and benchmarks~\cite{darkdiff2025}, reveal several key trends in the field of LLIE.
    Diffusion models, such as GlobalDiff~\cite{hou2023global} and DarkDiff, are demonstrating emerging dominance in perceptual quality, often achieving leading LPIPS scores. 
    For instance, DarkDiff reports state-of-the-art LPIPS on the challenging SID dataset, indicating its strength in generating perceptually realistic details even from noisy RAW inputs. 
    This aligns with the inherent capability of diffusion models to learn complex data distributions and generate diverse, plausible image content~\cite{he2024diffusion}.

    However, in terms of traditional metrics like PSNR and SSIM, highly optimized CNN-based methods like CIDNet~\cite{Verma_2025_CVPR} or Transformer-based models like Retinexformer~\cite{zamir2022restormer} often remain competitive or even lead, particularly on datasets like LOL and LOLv2.
    This is not entirely surprising, as these metrics favor pixel-wise fidelity to a single ground truth, which regression-based or structurally constrained models can excel at. The ``best'' method often depends on the specific dataset and the chosen evaluation metric, highlighting the absence of a single universally superior approach and underscoring the importance of comprehensive benchmarking across diverse data and metrics. For example, a model excelling on the LOL dataset, which predominantly features indoor scenes with specific noise characteristics, might not achieve the same rank on the SID dataset, which focuses on extreme low-light RAW images with severe noise.

    Qualitatively, diffusion models are generally praised for producing images with fewer visual artifacts compared to GANs, which can sometimes introduce unrealistic textures or color shifts. Diffusion models tend to generate smoother and more globally coherent results. However, if not well-conditioned or if accelerated too aggressively, they can sometimes produce overly smooth outputs or fail to recover very fine, non-repeating textures in extremely dark regions. Color accuracy can also be a challenge, as noted by the PyDiff authors who introduced a color corrector. Transformer-based models, with their strength in capturing global context, often yield good structural consistency but might require specific generative designs to match the pixel-level detail generation of top diffusion models or GANs.

    A critical battleground is the ``efficiency frontier.'' While diffusion models show excellent quality, their computational cost (high FLOPs, many parameters, slow inference) remains a major hurdle. In Table~\ref{tab:llie_sota_improved} illustrates this: models like GlobalDiff are powerful but computationally intensive. In contrast, methods like L2DM (latent diffusion), ReDDiT (distilled trajectory), and PyDiff (pyramid distillation) represent active efforts to push diffusion model performance while drastically reducing computational demands, making them more practical. Lightweight Transformers like IAT\cite{Cui_2022_BMVC} and LYT-Net~\cite{Brateanu2025} also showcase impressive efficiency. This ongoing research aims to find a better balance on the quality-complexity trade-off curve.

%%%%%%%%%%%%%%%%%%%%%%%%%%%%%%%%%%%
\section{Challenges and Limitations}
\label{sec:challenges_limitations}
%%%%%%%%%%%%%%%%%%%%%%%%%%%%%%%%%%%

    Despite the significant advancements brought by diffusion models to LLIE, several challenges and limitations persist, hindering their widespread adoption and optimal performance in all scenarios. These issues span computational demands, generalization capabilities, data dependencies, the balance between quality and performance, interpretability, and ethical considerations. Addressing these challenges is crucial for the continued progress of the field. A recent survey on diffusion models in low-level vision also highlights many of these general challenges, such as computational overhead and generalization~\cite{he2024diffusion}.

    \subsection{Computational Overhead and Inference Latency}
    The iterative denoising process, which is fundamental to the generative power of diffusion models, is also their primary source of computational expense. Requiring hundreds or even thousands of sequential neural network evaluations to generate a single enhanced image leads to high inference latency and significant GPU memory consumption. This ``scalability paradox", where models scale well in learning complex distributions for high-quality generation but poorly in computational efficiency for inference, severely restricts their deployment in real-time applications such as live video enhancement or on resource-constrained platforms like mobile phones and edge computing devices. For instance, the PyDiff paper explicitly notes that the constant resolution in standard diffusion reverse processes limits speed~\cite{pydiff2023}, and challenges in deploying complex models on mobile devices due to RAW image sizes and processing power are well-recognized. While accelerated diffusion techniques (Section~\ref{subsec:efficiency}) aim to mitigate this, they often involve a trade-off, potentially sacrificing some measure of enhancement quality or stability for speed.
    
    \subsection{Generalization and Robustness}
    Ensuring that LLIE diffusion models perform reliably across diverse and unseen lighting conditions, noise types, and scene contents remains a key challenge. Models often exhibit performance degradation when faced with out-of-distribution (OOD) inputs that differ significantly from their training data~\cite{zeroled2024}. Extreme darkness, highly non-uniform illumination, and sensor-specific noise patterns that were not part of the training manifold can lead to suboptimal enhancement or artifact generation. This is partly due to a ``prior mismatch'': diffusion models rely on learned priors about image characteristics. If a real-world low-light image presents features drastically different from these learned priors (\textit{e.g.}, unusual noise statistics or dynamic range), the model may struggle to enhance it correctly, potentially forcing the output towards its learned manifold in a way that introduces artifacts~\cite{zeroled2024}. While diffusion models are generally robust to some degree of noise, their adversarial robustness in the context of LLIE, i.e., susceptibility to specifically crafted perturbations designed to degrade enhancement quality or introduce malicious content, is an area requiring further investigation, though it's a general concern for diffusion models~\cite{betser2025whitenedcliplikelihoodsurrogate}.

    \subsection{Data Dependency and Scarcity}
    The performance of supervised diffusion models, like many deep learning approaches, is heavily dependent on the availability of large-scale, high-quality, diverse paired training datasets. However, acquiring perfectly aligned low-light/normal-light image pairs from a wide variety of real-world scenes, capturing diverse lighting conditions and sensor characteristics, is exceptionally difficult, time-consuming, and costly. This scarcity of ideal paired data limits the generalization capabilities of supervised models, often leading to overfitting to the specific characteristics of the available training sets. This creates a ``vicious cycle of data and realism'': the lack of diverse real-world paired data leads to models trained on synthetic or limited real data, which may not perform robustly on truly ``in-the-wild'' images; this, in turn, makes it harder to demonstrate broad utility and secure resources for more comprehensive data collection efforts. While unsupervised, self-supervised, and zero-shot diffusion models (as discussed in Section~\ref{subsec:autonomous})) aim to bypass the need for paired data, they face their own challenges in achieving high fidelity and avoiding unwanted artifacts without explicit ground-truth guidance or strong, reliable domain assumptions.
    Their performance can be less predictable, and they may struggle when the underlying assumptions of their learning paradigm are violated.

    \subsection{Perceptual Quality vs. Fidelity vs. Efficiency Trade-offs}
    Achieving an optimal balance between perceptual quality (how pleasing the image looks to a human), fidelity (how accurately it reconstructs a known ground truth, if available), and computational efficiency (inference speed and resource usage) is an ongoing challenge in LLIE. 
    As discussed in Section~\ref{subsec:metrics}, metrics like PSNR/SSIM that measure fidelity may not align with human perception of quality, especially for generative models like diffusion methods that might produce plausible and visually appealing details not present in a specific reference image. 
    Diffusion models in LLIE can sometimes exhibit specific artifacts if not carefully designed or trained. These may include color shifts (a concern addressed in PyDiff~\cite{pydiff2023} and DarkDiff~\cite{darkdiff2025}), excessive smoothing leading to loss of fine textures (especially if the model is too conservative in extremely dark regions or if acceleration techniques are too aggressive), or a reduction in the diversity of generated details compared to the true underlying distribution if the model is over-regularized.
    The subjective nature of ``enhancement'' further complicates this: what one user considers a good enhancement (\textit{e.g.}, maximum brightness) another might find unnatural. 
    While guided diffusion models (Section~\ref{subsec:adaptive}) offer potential for tunable enhancement, defining the universally ``optimal'' guidance or balancing these conflicting objectives remains a non-trivial problem.
    
    \subsection{Interpretability and Explainable AI (XAI)}
    Like most deep neural networks, diffusion models operate largely as ``black boxes.'' Understanding precisely why a diffusion model produces a particular enhanced output, or why it generates a specific artifact, is difficult~\cite{zheng2022low}.
    This lack of interpretability can hinder trust in these models, especially when deployed in safety-critical applications such as autonomous driving or medical imaging (as alluded to in the abstract ). It also makes the process of debugging failure cases and systematically improving model behavior more challenging. Developing XAI techniques tailored for diffusion-based LLIE could not only build confidence but also provide insights into the model's internal representations and decision-making processes. Such understanding could, in turn, lead to more fine-grained and intuitive control mechanisms for the enhancement process, moving beyond current guidance signals to allow for more predictable and reliable manipulation of image characteristics.
    
    \subsection{Ethical Considerations and Responsible AI}
    The deployment of powerful generative models like diffusion models for LLIE also raises several ethical considerations that warrant careful attention.
    \begin{enumerate}
        \item \textbf{Bias Amplification:} If the training data contains biases related to, for example, skin tones, demographic representations, or types of scenes, the diffusion model may learn and even amplify these biases. This could lead to inequitable performance, where the model enhances images better for certain groups or scenes than for others.
        \item \textbf{Misuse and Malicious Applications:} While LLIE is primarily a restorative task, the underlying generative capabilities of diffusion models could potentially be misused. For instance, enhanced low-light images might be used to misrepresent events or, in a broader context of generative AI, to create more convincing deepfakes or manipulated media if the technology is adapted for such purposes.
        \item \textbf{Fairness in Enhancement Quality:} It is important to assess whether the LLIE model provides fair and consistent enhancement quality across different types of input images, including those featuring diverse demographic groups, varied environments, and different levels of initial degradation.
        \item \textbf{Environmental Impact:} Training very large-scale diffusion models incurs a significant computational cost, which translates to energy consumption and an environmental footprint. This is a broader concern for large AI models but is relevant as LLIE models grow in complexity.
        \item \textbf{Potential for ``}\texttt{Reality Distortion}\textbf{'':} A nuanced concern is that while LLIE aims to reveal information obscured by darkness, the powerful generative nature of diffusion models means they might also create plausible-seeming but ultimately false details, especially in regions of extreme degradation where true scene information is entirely absent. If such generated details are presented as factual recovery of information, this could have serious ethical implications in contexts like forensic analysis or surveillance, where the authenticity and reliability of visual evidence are paramount.
    Addressing these challenges requires ongoing research, careful dataset curation, robust evaluation protocols, and a commitment to developing and deploying AI technologies responsibly.
    \end{enumerate}

%%%%%%%%%%%%%%%%%%%%%%%%%%%
\section{Future Directions}
\label{sec:future}
%%%%%%%%%%%%%%%%%%%%%%%%%%%

    The field of diffusion models for LLIE is rapidly evolving, with numerous promising avenues for future research. Building upon the current taxonomy and addressing the identified challenges, several key directions emerge that could lead to more robust, efficient, controllable, and practical LLIE systems. These directions often involve integrating insights from different categories within the taxonomy or leveraging broader advancements in AI.

    \subsection{Harnessing Foundation Models for LLIE}
    The remarkable success of large-scale pre-trained models, or foundation models, in various domains presents a significant opportunity for LLIE.
    \begin{enumerate}[leftmargin=18pt,itemsep=0.3pt]
        \item \textbf{Adapting Pre-trained Diffusion Models:} Future research will likely focus on effectively adapting existing powerful image generation diffusion models (\textit{e.g.}, Stable Diffusion~\cite{stable_diffusion}, Imagen~\cite{imagenteamgoogle2024imagen3}) for the LLIE task. This could involve specialized fine-tuning techniques, novel prompting strategies (if text-to-image models are used), or zero-shot guidance mechanisms that steer the generation process towards enhanced low-light outputs without extensive LLIE-specific retraining~\cite{darkdiff2025}. The key challenge is to leverage the rich visual priors of these foundation models while ensuring fidelity to the low-light input and appropriate enhancement.
        \item \textbf{Multi-modal Foundation Models:} Exploring multi-modal foundation models that can process and integrate information from text, audio, or other sensors alongside visual data could lead to more contextually aware and controllable LLIE. For instance, a textual description of desired lighting conditions or objects of interest could guide the enhancement.
    \end{enumerate}

    \subsection{Towards Real-Time and On-Device LLIE with Diffusion Models}
    Addressing the computational bottleneck of diffusion models is paramount for their practical application, especially in real-time video enhancement and deployment on resource-constrained mobile or edge devices. 
    \begin{enumerate}[leftmargin=18pt,itemsep=0.3pt]
        \item \textbf{Advanced Acceleration Techniques:} Continued research into model compression (pruning, quantization), knowledge distillation, and the design of more efficient network architectures specifically for diffusion-based LLIE is crucial. This includes developing novel sampling algorithms that can drastically reduce the number of required denoising steps with minimal degradation in visual quality.
        \item \textbf{Hardware Co-design:} The symbiotic relationship between AI models and hardware suggests that future progress may also come from hardware specialized for diffusion model computations, or diffusion models designed to better leverage existing mobile/edge NPUs.
    \end{enumerate}

    \subsection{Principled Unsupervised, Self-Supervised, and Zero-Shot Learning}
    Given the difficulty of obtaining diverse, high-quality paired data for LLIE, autonomous learning paradigms remain a vital research direction.
    \begin{enumerate}[leftmargin=18pt,itemsep=0.3pt]
        \item \textbf{Robust Degradation Modeling:} Developing more sophisticated unsupervised methods that can learn better representations of real-world low-light degradations is essential. This includes more accurately modeling the complex, non-uniform, and signal-dependent noise characteristics unique to low-light photography. By learning the degradation process itself, models could perform enhancement without explicit ground-truth supervision, improving generalization to unseen camera sensors and environments.
        \item \textbf{Domain Generalization and Adaptation:} A key frontier is creating models that robustly generalize across different domains (e.g., from indoor synthetic data to outdoor night scenes) without requiring retraining. Future work should focus on unsupervised domain adaptation techniques that allow a diffusion model trained on one dataset to maintain high performance on a completely different target domain, addressing the critical challenge of data scarcity and dataset bias.
        \item \textbf{Disentangled Representation Learning:} Future unsupervised models could aim to learn disentangled representations of content, illumination, noise, and color. By separating these factors, a model could offer more principled and controllable enhancement, allowing a user to, for instance, adjust illumination without altering the learned noise characteristics or color palette, moving beyond simple end-to-end mappings to allow for more predictable and reliable manipulation of image characteristics.
    \end{enumerate}

    \subsection{Enhancing Controllability and Interpretability}
    As LLIE models are deployed in higher-stakes applications, ensuring their outputs are controllable and their decision-making processes are understandable is paramount.
    \begin{enumerate}[leftmargin=18pt,itemsep=0.3pt]
        \item \textbf{Fine-Grained Semantic Control:} Building on the progress in Guided Diffusion, future research should move beyond simple text prompts or masks towards more fine-grained, semantic control. This could involve instruction-based models like InstructIR that understand nuanced commands (e.g., ``brighten the person's face but keep the background moody'') or models that can be guided by example images to match a desired artistic style or lighting effect.
        \item \textbf{Explainable AI (XAI) for LLIE:} The ``black box'' nature of diffusion models hinders trust and makes debugging difficult. A critical direction is the development of XAI techniques tailored for generative LLIE. This could include methods for visualizing which parts of the low-light input most influenced the final enhanced output, or attributing the generation of specific details or artifacts to certain steps in the reverse diffusion process.
        \item \textbf{Hybrid Models for Principled Generation:} The most robust future systems will likely be hybrid models that synthesize strengths from across our taxonomy. For instance, a model could leverage a physically-grounded prior from the Intrinsic Decomposition category to guide the overall enhancement trajectory, while using a powerful generative diffusion process from the Autonomous category to plausibly synthesize fine-grained textures. This entire process could then be made efficient by operating in a compressed representation from the Spectral \& Latent category, creating a system that is principled, powerful, and practical.
    \end{enumerate}

%%%%%%%%%%%%%%%%%%%%%%%%%%%
\section{Conclusion}
\label{sec:conclusion}
%%%%%%%%%%%%%%%%%%%%%%%%%%%
    This survey presents the first comprehensive and critical review of diffusion models for low-light image enhancement (LLIE), offering a structured taxonomy that categorizes over 30 recent methods into six key perspectives: Intrinsic Decomposition, Spectral \& Latent, Accelerated, Guided, Multimodal, and Autonomous diffusion models. By systematically analyzing these approaches across methodological, computational, and application-driven axes, we identify core trade-offs that define the current state of the field.

    Our survey reveals three defining trends. \textit{First}, the performance landscape is governed by an efficiency-fidelity trade-off, where high-quality restoration often comes at the expense of inference speed. This motivates the development of accelerated diffusion techniques, including trajectory optimization, latent-space operations, and knowledge distillation. \textit{Second}, the field is undergoing a shift from monolithic restoration to controllable, task-aware enhancement, as seen in the rise of guided and multimodal diffusion frameworks that integrate semantic cues, user instructions, and sensor fusion. \textit{Third}, the need for greater generalization in real-world settings has spurred the growth of autonomous diffusion models that minimize reliance on paired data through unsupervised, zero-shot, and domain-adaptive learning strategies.
    
    Despite significant progress, practical deployment remains hindered by challenges such as computational overhead, interpretability, domain mismatch, and the lack of robust metrics that align with human perception and downstream utility. Future work will require bridging these gaps through hybrid strategies that combine physical priors, data-driven learning, and foundation model guidance—while ensuring ethical, fair, and energy-efficient deployment.
    
    By mapping the design space, benchmarking performance, and surfacing open challenges, this survey aims to serve as a foundation for the next generation of diffusion-based LLIE research and applications.

\bibliographystyle{unsrt}
\bibliography{reference}

\clearpage\newpage
\appendix

\end{document}